\documentclass[10pt,twocolumn,letterpaper]{article}

\usepackage{iccv}
\usepackage{times}
\usepackage{epsfig}
\usepackage{graphicx}
\usepackage{amsmath}
\usepackage{amssymb}
\usepackage{amsmath}
\usepackage{amssymb}
\usepackage{mathtools}
\usepackage{amsthm}
\usepackage{url}
\usepackage{subcaption}
\usepackage{booktabs} % for professional tables
\usepackage{multirow}
\usepackage{algorithm}
\usepackage{algorithmic}
\usepackage{bbding}
\usepackage{pifont}
\usepackage{footnote}
\usepackage{overpic}
\usepackage{comment}
\usepackage{amssymb}
\usepackage{threeparttable}
\usepackage[pagebackref=true,breaklinks=true,letterpaper=true,colorlinks,bookmarks=false]{hyperref}

\usepackage{makecell}
\usepackage{sidecap}

\DeclareMathOperator*{\argmax}{arg\,max}
\DeclareMathOperator*{\argmin}{arg\,min}

\usepackage[capitalize,noabbrev]{cleveref}

\theoremstyle{plain}

\theoremstyle{definition}

\theoremstyle{remark}

\iccvfinalcopy % *** Uncomment this line for the final submission

\def\iccvPaperID{2069} % *** Enter the ICCV Paper ID here

\ificcvfinal\fi

\begin{document}

%%%%%%%%% TITLE
\newcommand{\nameofmethod}{{{dataset quantization}}}
\title{Dataset Quantization}

\author{
Daquan Zhou \textsuperscript{1}\footnotemark[1] 
\quad Kai Wang\textsuperscript{2}\thanks{Equal first author.} 
\quad Jianyang Gu\textsuperscript{2}\footnotemark[1]
\quad Xiangyu Peng\textsuperscript{2}
\quad Dongze Lian\textsuperscript{2}\\
\quad Yifan Zhang\textsuperscript{2}
\quad Yang You\textsuperscript{2}\footnotemark[2]
\quad Jiashi Feng\textsuperscript{1}\thanks{Corresponding author.}
\\
\textsuperscript{1}{Bytedance Inc.} \quad
\textsuperscript{2}{National University of Singapore}
\\
\small{\texttt{zhoudaquan21@gmail.com}} \quad \small{\texttt{kai.wang@comp.nus.edu.sg}} \quad
\small{\texttt{gu\_jianyang@zju.edu.cn}} \\
\quad \small{\texttt{youy@comp.nus.edu.sg}} \quad \small{\texttt{jshfeng@bytedance.com}}
\\
\small{Code: \url{https://github.com/magic-research/Dataset_Quantization}}
}

\maketitle
\ificcvfinal\thispagestyle{empty}\fi

%%%%%%%%% ABSTRACT
\begin{abstract}
   State-of-the-art deep neural networks are  trained  with large amounts (millions or even billions) of data. The expensive computation and memory costs make it  difficult to train them on limited hardware resources, especially for recent popular large language models (LLM) and computer vision models (CV). Recent popular dataset distillation methods are thus developed, aiming to reduce the number of training samples via synthesizing small-scale datasets via gradient matching.
   However, as the gradient calculation is coupled with the specific network architecture,
   the synthesized dataset is  biased and performs poorly when used for training unseen architectures. To address these limitations, we present \emph{dataset quantization} (DQ), a new framework to compress large-scale datasets into small subsets which can be used for training any neural network architectures.
Extensive experiments demonstrate that DQ is able to generate condensed small datasets  for training unseen network architectures with state-of-the-art compression ratios for lossless model training. 
To the best of our knowledge, DQ is the first method that can successfully distill   large-scale datasets such as ImageNet-1k with a state-of-the-art compression ratio.
Notably, with 60\% data from ImageNet and 20\% data from Alpaca's instruction tuning data, the models can be trained with negligible or no performance drop for both vision tasks (including classification, semantic segmentation, and object detection) as well as language tasks (including instruction tuning tasks such as BBH and DROP). 
\end{abstract}

\begin{figure}[tp]
    \centering
    \includegraphics[width=0.5\textwidth]{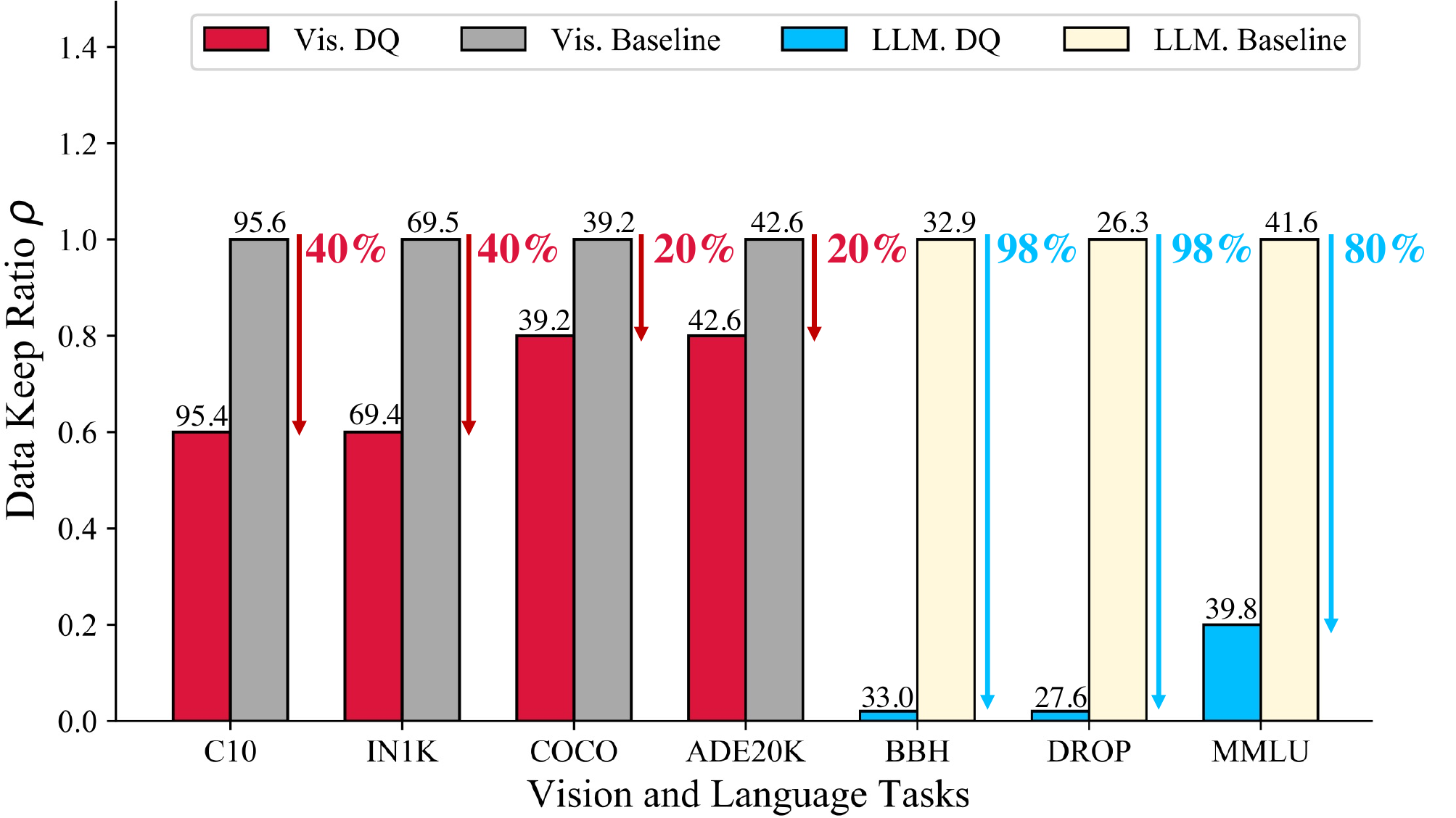}
    \caption{\textbf{Lossless dataset compression with Dataset Quantization (DQ) framework.} On both vision and language tasks. In the plot, we use ResNet18 as backbone for all tasks and LLaMA-7B for all language tasks with instruction fine-tuning. 
    }
    \label{fig:fig1}
\end{figure}

\section{Introduction}

Deep neural networks have shown superior performance in a wide range of fields such as computer vision~\cite{he2016deep, he2017mask, dosovitskiy2020image,zhang2022expanding} and natural language processing \cite{devlin2018bert, bao2021beit}. 
Their performance  depends heavily on the amount of training data. For example, recent state-of-the-art models \cite{kolesnikov2020big,wang2022image,dean2021introducing, yu2022coca} on ImageNet-1K takes three billion data for pre-training. 
This is hardly affordable for researchers with limited computational resources.
However, are all the data in the large dataset beneficial or necessary to the training? Is it possible to remove some redundant samples without degrading the training performance? What is the performance of the pretrained models with less data on downstream tasks? In this paper, we conduct extensive experiments and conduct detailed explorations on those questions. 
To address the first question,  several Dataset Distillation (DD) algorithms~\cite{zhao2021DC,zhao2021dsa,kim2022dataset,zhaodm,wang2022cafe,cazenavette2022distillation,ftd, wang2023dim, liu2023dream} are proposed recently to reduce the training dataset size by synthesizing a new set of data that is significantly smaller than the original one. With the new synthesized dataset, the training cost is reduced significantly, while yielding comparable results with the models trained on the original datasets.

\begin{figure*}[t]
    \centering
    \begin{subfigure}[]{0.32\textwidth}
    \begin{overpic}[width=\textwidth, height=0.86\textwidth]{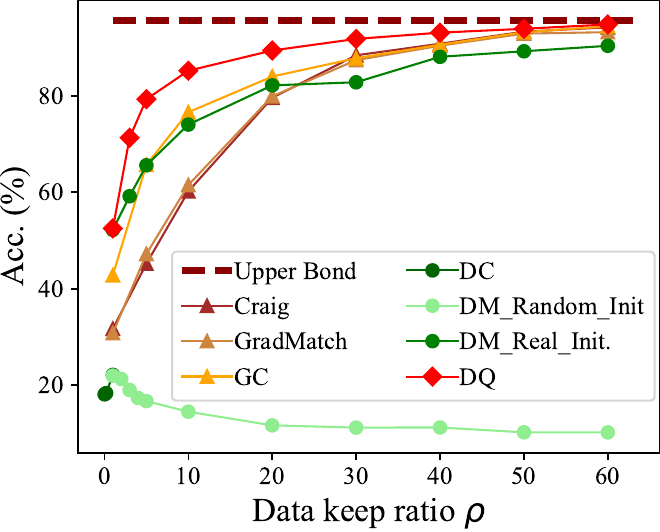}
    \put(44,57){
    \tiny
    \setlength\tabcolsep{1mm}
    \renewcommand{\arraystretch}{1}
    \begin{tabular}{c|ccc} 
    \toprule
      Method  & Hours & IN-1K  & COCO  \\ \midrule
      DM-60  &  28K &- &-\\
      DQ-60  &  72 &89.0 &39.0\\
      Full  &-  &89.4 &39.2\\
      \bottomrule
    \end{tabular}}
    \end{overpic}
    \caption{}
    \label{fig:1a}
    \end{subfigure}
    \raisebox{0.16cm}{
    \begin{subfigure}[]{0.23\textwidth}
        \includegraphics[width=\textwidth, height=1.25\textwidth]{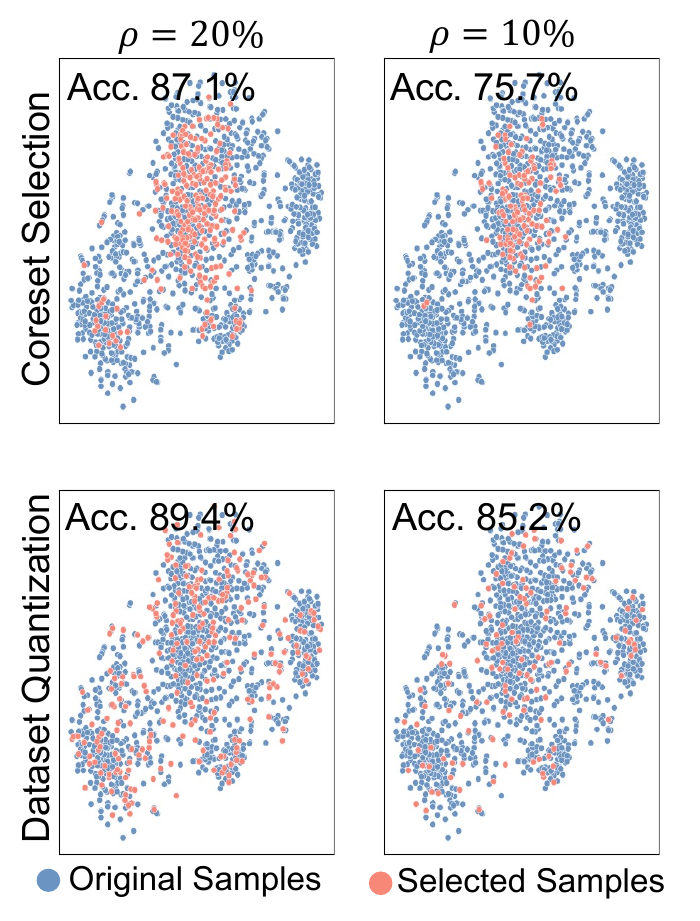}
        \caption{}
        \label{fig:1b}
    \end{subfigure}}
    \raisebox{0.16cm}{
    \begin{subfigure}[]{0.39\textwidth}
        \includegraphics[width=\textwidth, height=0.75\textwidth]{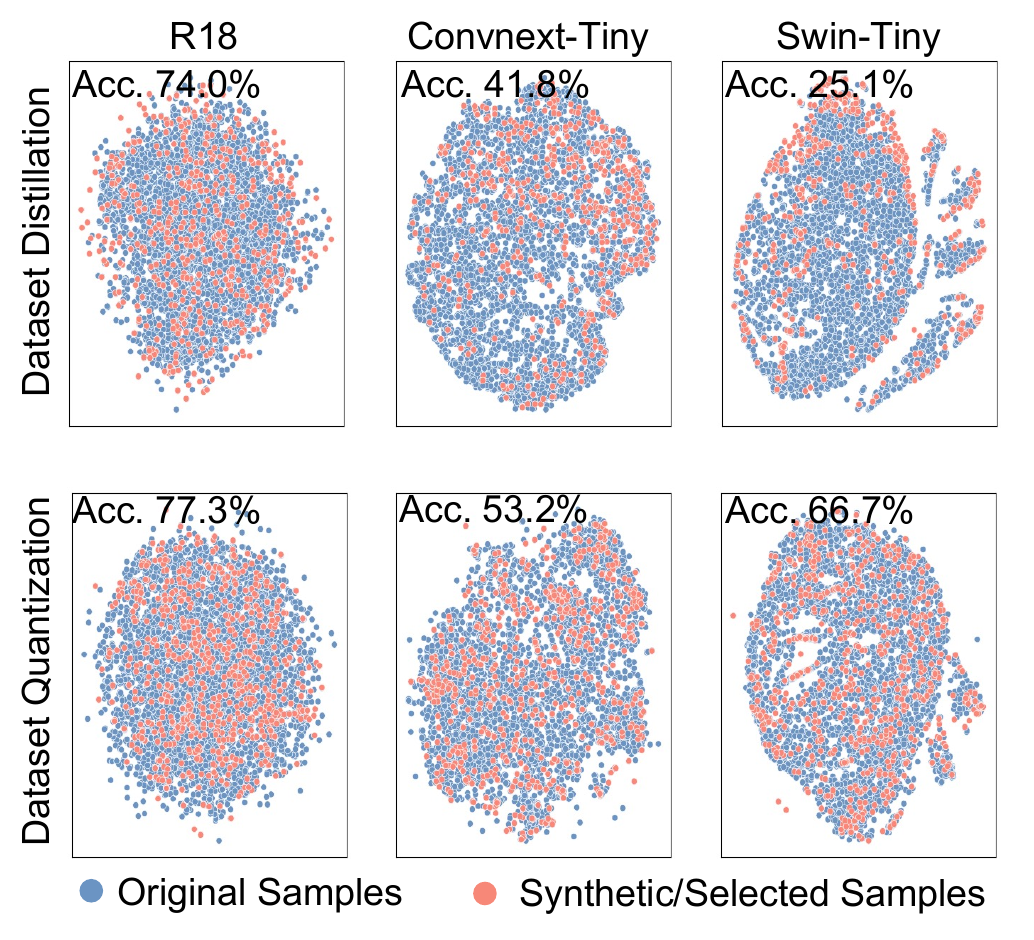}
        \caption{}
        \label{fig:1c}
    \end{subfigure}}

\caption{\textbf{Our proposed dataset quantization outperforms existing  dataset distillation and coreset selection methods significantly}. 
(a)  Model training accuracy from DD (DC~\cite{zhao2021DC} and DM~\cite{zhaodm}), coreset selection (Craig~\cite{mirzasoleiman2020coresets}, GradMatch~\cite{killamsetty2021glister}, and GC~\cite{iyer2021submodular}), and our proposed DQ across different data keep ratios.
`Hours' denotes the time for compressing ImageNet dataset with 60\% data keep ratio.
(b) Visualization of the samples diversity of GraphCut and DQ, where $\rho$ is the data keep ratio (better in color).  
(c)  Cross-architecture visualization of the feature distributions among the dataset generated by a dataset distillation methods `distribution matching' (DM) and DQ on ResNet-18 on CIFAR-10 bird class. Compared with DM, our proposed DQ effectively captures the whole dataset distribution for all the architectures, thus generalizing better.
}
\end{figure*}

Although having made significant progress, two limitations make those algorithms hard to be deployed in an industrial environment:
i) \textbf{Poor generalization capability.} They all rely on specific metrics to match the synthetic and real samples \cite{zhao2021dataset,zhaodm}. Thus the synthetic datasets are inevitably biased by the model architecture   involved in the metric computation, resulting in poor performance when used for training unseen model architectures. For example, as shown in Fig.~\ref{fig:1c}, \textbf{the dataset synthesized based on ResNet-18 \cite{he2016deep} suffers a 59.4\% accuracy drop when used for training  Swin-Tiny \cite{liu2021swin} (81.2\% vs 21.6\%)}. 
ii) \textbf{Low scalability to larger datasets.} Different from other deep learning tasks that optimize the parameters of a given architecture, dataset distillation aims to optimize the synthetic set,
the computational cost is quadratically proportional to the size of the synthetic set. When the size is large, the computational cost becomes unaffordable. For example, as in Fig.~\ref{fig:1a}, previous SOTA method DM~\cite{zhaodm}
needs \textbf{28, 000 GPU hours to distill ImageNet-1K with 60\% data processing.}

To address these limitations, we explore a different direction from synthesizing samples based on our empirical observations that the samples selected by coreset methods~\cite{iyer2013submodular,guo2022deepcore, chen2010super, agarwal2020contextual, qin2023infobatch}
could be used to train unseen network architectures (\textit{i.e.} good cross-architecture generalization).
However, as the data keep ratio is small, the selected samples tend to lose the diversity, leading to a low performance for model training. As in the first row in Figure \ref{fig:1b}, coreset methods tend to sample data points in a biased region. This led to a significant accuracy drop when used for model training. As shown in Fig. 6 in the following section, our proposed DQ is able to achieve 10\% (75.7\% vs 85.2\%) higher accuracy over the previous SOTA coreset method.

In this paper, we aim to develop a method that combines the advantages of Dataset Distillation methods and  the Coreset methods: a unified dataset compression method that   generates compact datasets useful for training various network architectures while maintaining state-of-the-art training performance under all data keep ratios.
We start with investigating the reason behind the poor performance of the coreset selection method \cite{shim2021core}  under low data keep ratio, and we find it lies in   the one-time selection strategy, resulting in a low diversity of the selected data. 
This will lead to a significant performance drop as shown in Fig. \ref{fig:1b}. More detailed analysis on previous coreset selection methods~\cite{iyer2021submodular, killamsetty2021glister} can be found in Sec. \ref{subsec:extractor} and in the Appendix.

We thus propose a new pipeline to overcome the aforementioned issues of the coreset algorithm and term it Dataset Quantization (DQ).
Specifically, 
DQ first divides the entire dataset into a set of non-overlapping bins recursively based on the submodular gains~\cite{iyer2021submodular} that aims to maximize the diversity gains as defined in Eqn.~\ref{eqn:graphcut}.
Then, a small portion of data samples is uniformly sampled from all bins. In this manner, the selected samples are optimized to cover as much as possible the entire dataset with the inter-data diversity maximized.
We prove mathematically that the dataset selected by DQ indeed has larger diversity than the coreset selection based methods.
Motivated by recent patch-based image representation \cite{dosovitskiy2020image, he2022masked, zhou2022understanding}, we measure the importance scores of patches and save the most important ones to reduce the storage cost. At the training stage, we reconstruct training images via important patches and a pre-trained MAE~\cite{he2022masked} model.

Different from dataset distillation methods, as shown in the second row in Fig.~\ref{fig:1c},
the quantized dataset maintains a high coverage  over the entire data in the latent feature space
across different model architectures. The validation accuracy is also significantly higher than those models trained with DD algorithms (\eg, 34.4\% higher for ViT-Tiny).
Compared with DD methods, \textbf{DQ only takes 72 GPU hours to quantize ImageNet data with 60\% keep ratio, which is 388$\times$ (28, 000 vs 72 GPU hours) faster}, while achieving much higher performance on large data keep ratios. 
On the other hand, when comparing to coreset selection methods, as shown in Fig. \ref{fig:1a} and \ref{fig:1b}, DQ selects samples with larger diversity and achieves better performance when the data keep ratio is low (10\% data kept).

We conduct extensive experiments and show that the proposed \nameofmethod~method is able to generate compact datasets that can be used to train unseen models such as model families from ViT, ResNet and MobileNetV2, LlaMA, etc. 

Specifically, 
\textbf{for vision tasks}, on CIFAR-10 and ImageNet-1K, only 60\%  of the data are used to train the models to achieve a comparable model performance as those trained with full datasets.
\textbf{for language tasks}, on BBH and DROP benchmark, only 2\% instruction data are needed to achieve comparable model performance as those trained with full datasets.
We further verify that the model weights pre-trained on the quantized dataset can be generalized into downstream tasks such as object detection and segmentation.
As shown in Fig.~\ref{fig:downstream}, the ResNet-50 \cite{he2016deep} model pre-trained on 60\% ImageNet also achieves negligible performance drop when finetuned on COCO~\cite{lin2014microsoft} (39.0\% vs 39.2\%) and ADE20K~\cite{zhou2019semantic} (42.3\% vs 42.5\%). 

Our main contributions are summarized as:
\begin{itemize}
  \item We propose a new framework, Dataset Quantization (DQ), to compress datasets into a small compact one that can be used for training unseen network architectures with state-of-the-art compression performance. 
  
  \item We propose a scalable and efficient dataset compression algorithm that can be used for large dataset such as ImageNet-1K. With Dataset Quantization, we are able to remove 40\% data from ImageNet-1K dataset and 80\% data from the Alpaca instruction dataset with no training performance loss. 
  
  \item We verify that the models trained with a compressed dataset can be used for downstream tasks. The models pre-trained with 60\% of data on ImageNet-1K achieve no performance on COCO for object detection and ADE20K for segmentation.
\end{itemize}

\label{intro}

\section{Related work}
\label{sec:related}
In this section, we review two representative  related methods: dataset distillation and coreset selection. We also introduce limitations and analysis of these two kinds of methods.
\subsection{Dataset distillation}
 Dataset distillation (DD)~\cite{wang2018dataset} is the first method that proposes to synthesize a small amount of informative samples from a large dataset. 
 Specifically, it optimizes the synthetic samples by minimizing the   loss on the original training samples of the models trained on the synthetic ones. 
 Afterwards, a series of  techniques have been proposed such as Dataset condensation (DC)~\cite{zhao2021dataset}, DSA~\cite{zhao2021dsa} and IDC~\cite{kimICML22}. These methods propose to match the loss gradient calculated from the original and   synthetic data. 
CAFE ~\cite{wang2022cafe} and DM ~\cite{zhaodm} introduce a feature distribution matching strategy to reduce the potential bias from large-gradient samples. 
A recent work~\cite{cazenavette2022distillation} tries to minimize the difference of training trajectories between original and synthetic samples. 

\subsection{Coreset selection}
\label{coreset_related}
 Coreset selection has been actively explored     for compressing datasets, which aims  to select a subset of the most representative samples out of the target dataset.
The previous methods have proposed  different selection criteria: geometry-based~\cite{chen2010super, agarwal2020contextual, sener2018active, sinha2020small}, uncertainty-based~\cite{coleman2019selection}, error-based~\cite{toneva2018empirical, paul2021deep}, decision-boundary-based~\cite{ducoffe2018adversarial,margatina2021active},  gradient-matching~\cite{mirzasoleiman2020coresets, pmlr-v139-killamsetty21a}, bilevel optimization~\cite{killamsetty2021glister} and submodularity-based methods~\cite{iyer2021submodular}. 
Among them, the Contextual Diversity (CD)~\cite{agarwal2020contextual}, Herding~\cite{welling2009herding}, and k-Center Greedy~\cite{sener2018active} try to remove the redundant samples based on their similarity to the remaining  samples.
Cal~\cite{margatina2021active} and Deepfool~\cite{ducoffe2018adversarial} argue that the coreset should be selected based on their difficulties for learning.
Craig ~\cite{mirzasoleiman2020coresets} and GradMatch~\cite{pmlr-v139-killamsetty21a} try to find an optimal coreset that has the similar gradient values with the whole dataset when training them on a network.
Glister~\cite{killamsetty2021glister} introduce a validation set to maximize the log-likelihood with the whole dataset, where involves a time-consuming bilevel optimization.
FL~\cite{iyer2021submodular} and Graph Cut (GC)~\cite{iyer2021submodular} consider the diversity and information simultaneously. 
\begin{table}[t]
    \centering
    \footnotesize
    \caption{Comparisons of the Dataset Distillation (DD), Coreset selection and our proposed Dataset Quantization (DQ). DQ combines the advantages of DD and coreset selection and is better at compressing datasets for training modern deep neural networks. }
    \begin{tabular}{l|c c c c c}
    \toprule
         \multirow{2}{*}{Method}  & Arch. & \multirow{2}{*}{Scalable} & Time  & \multirow{2}{*}{Diverse} & Data  \\
         & generalized & & Efficient & & Efficient \\
         \midrule
         DD & \ding{55}  & \ding{55}  & \ding{55} & \ding{51} &\ding{51} \\
          Coreset  & \ding{51}  & \ding{51}  & \ding{51} &  \ding{55}  & \ding{55} \\
          DQ  & \ding{51}  & \ding{51}   & \ding{51}  &  \ding{51}   & \ding{51} \\
    \bottomrule
    \end{tabular}
    \label{tab:related_comp}
\end{table}

\subsection{Limitations and analysis}
DD methods are hard to be applied on large datasets or architectures, such as ImageNet-1K or ResNet series, mainly due to the following limitations:
(i) \textit{Poor generalizability}. As shown in Fig. \ref{fig:1b}, 
the synthesized images only work well on the same model architecture providing the optimization supervision, 
while fail training on other model architectures. 
(ii) \textit{Poor scalability}. As the green line shows in Fig. \ref{fig:1a}, they saturate fast as the data keep ratio increases and can never reach the performance of the original datasets. 
(iii) \textit{High  computational cost} for large datasets. As shown in the mini table in Fig.~\ref{fig:1a}, compressing the whole ImageNet into 60\% subset requires 28K GPU hours in total.  

The above shortcomings are overcame  by the coreset selection methods. 
However, the diversity  of the coreset   samples is not guaranteed under low data keep ratio, leading to worse  performance than DD methods at low-data regime~\cite{zhao2021dataset}, as shown in Fig.~\ref{fig:1a} and Fig.~\ref{fig:3a}. 
Tab. \ref{tab:related_comp} summarizes the differences among DD, coreset selection and DQ. Across all the five aspects,   our proposed dataset quantization method consistently performs better.

\label{related}

\section{Method}
As mentioned in Sec.~\ref{sec:related}, the synthetic dataset based on DD methods performs poorly for training unseen network architectures as the matching metrics are coupled with the utilized network. We are thus motivated to explore a data selection strategy that is not sensitive to model architectures. In this section, we first introduce preliminaries about the  coreset selection method and theoretically analyze its limitation. In particular, we choose the GraphCut based method~\cite{iyer2021submodular} as an example.   Then, we present details of our proposed dataset quantization (DQ) method.

\begin{figure*}[htp]
    \centering
    \includegraphics[width=1.0\textwidth]{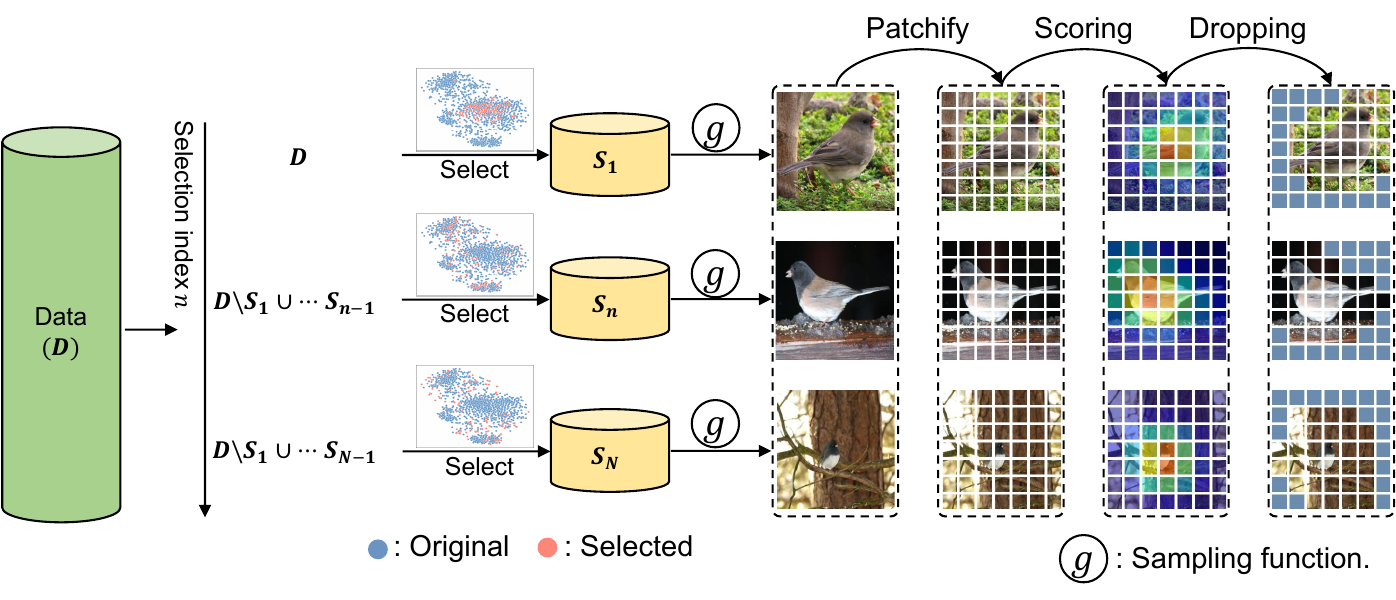}
    \caption{An overview of the proposed DQ framework. We first divide the whole dataset $\mathbf{D}$ into $N$ non-overlapping bins $\mathbf{S}_n$. Then, the $\mathbf{S}^*$ is  aggregated from $N$ bins by a sampling function.
    After that, to further reduce the redundancy from each image, DQ drops a fraction of patches with the lowest information and reconstruct samples at the training stage via MAE. 
    }
    \label{fig:ppl}
\end{figure*}

\subsection{Preliminary of coreset selection}
\label{subsec:extractor}

Coreset-based algorithms~\cite{chen2009coresets,shim2021core,huang2020coresets} address the limitations of DD methods. However, almost all coreset selection methods only select a single subset from the entire dataset in a one-stop manner. We empirically observe that it inevitably introduces severe \textit{selection bias}\textemdash the samples lying in the high-density regions of the dataset distribution are more often selected than others\textemdash and yields selection results with limited variety. 
We provide more detailed theoretical analysis for the observation. 

\textbf{Theoretical Analysis for Coreset Selection.}
As mentioned in Sec.~\ref{coreset_related}, almost all coreset selection methods utilize a heuristic metric to select samples, which is hard to avoid selecting some samples that have similar performances under the heuristic metric. GraphCut~\cite{iyer2021submodular}, a recent state-of-the-art method, we choose it as an example to analyze the coreset selection process.
$\mathbf{D} = \{(x_k, y_k)\}_{k=1}^M$ denotes $M$ labeled samples. We default to select $K$ samples from $\mathbf{D}$ to form a coreset. The coreset is initialized as $\mathbf{S}_{1}^{1} \leftarrow \emptyset$ and updated as $\mathbf{S}_{1}^{k} \leftarrow \mathbf{S}_{1}^{k-1} \cup x_{k}$.
Note that, $\mathbf{S}_n$ denotes $n-$th bin, $\mathbf{S}_{n}^{k}$ represents first $k$ samples of $n-$th bin and $x_{k}$ is the $k-$th selected sample.
We define the feature extractor as $f(\cdot)$.
In GraphCut, samples are selected via maximizing submodular gains~\cite{iyer2021submodular} $P(x_k)$ in feature space, which is defined as follows,
\begin{equation}
\label{eqn:graphcut}
\small
    P(x_{k}) = \sum_{p \in \mathbf{S}_{1}^{k-1}}\underbrace{||f(p)-f(x_{k})||_{2}^{2}}_{C_{1}(x_{k})} - \sum_{p \in \mathbf{D} \backslash \mathbf{S}_{1}^{k-1}}\underbrace{||f(p)-f(x_{k})||_{2}^{2}}_{C_{2}(x_{k})},
\end{equation}
where $\mathbf{S}_{1}^{k-1}$ denotes the set of selected samples and $\mathbf{D} \backslash \mathbf{S}_{1}^{k-1}$ represents the remained set. GraphCut aims to maximize $P(x_{k})$: it expects to maximize the diversity between $x_{k}$ and the selected set while minimizes the distance between $x_{k}$ and the remained set. Thereby $\mathbf{S}_1$ is expected to be a coreset covering the original distribution while maintaining largest diversity. However, as $K \ll M$, the sum value of $C_1(x_{k})$ is far smaller than $C_2(x_{k})$. 
The distance between $x_k$ and the remained set takes the dominant position in the gain calculation. 
Thus the diversity of selected $K$ samples is not guaranteed as expected. Especially when the data keep ratio is low.

Mathmatically, supposing the average feature is at the origin,
we define the maximum radius of set $\mathbf{S}_1^{k-1}$ as $\mathbf{R}_{1}^{k-1} = \max_{p \in \mathbf{S}_{1}^{k-1}}||f(p)||_{2}$, we prove the continuous solution of the next selected sample $x_{k}$ needs to satisfy 
\begin{equation}
\label{eqn:solution}
    ||f(x_{k})||_{2}^{2} \leq (\frac{2k}{M-2k})^{2}(\mathbf{R}_{1}^{k-1})^{2}.
\end{equation}
As $M \gg k$, the exact solution of $f(x_{k})$ is within $(\mathbf{R}_{1}^{k-1})^{2}$ or as an outlier point that is as close as possible to the boundary of ball $\mathbf{R}_{1}^{k-1}$.
The theoretical analysis well aligns the visualization in Fig. \ref{fig:1b}. The diversity of selected samples is hard to be guaranteed for coreset selection. We provide more detailed proof in Appendix.

From the above analysis, 
the main reason of poor coreset diversity of GraphCut is $M \gg k$ (\textit{i.e.} over-large denominator) in Eq.~\eqref{eqn:solution}. 
There naturally rises an idea of recursively selecting from $\mathbf{D}$ for several times.
Assume that we select $\mathbf{S_{2}}$ from dataset $\mathbf{D} \backslash \mathbf{S_{1}}$ again. The maximum radius of set $\mathbf{S_{2}^{k-1}}$ can be denoted as $\mathbf{R}_{2}^{k-1} = \max_{p \in \mathbf{S}_{2}^{k-1}}||f(p)||_{2}$. As the most compact subset has been selected in $\mathbf{S_{1}}$, $\mathbf{R}_{2}^{k-1}$ is obviously larger than $\mathbf{R}_{1}^{k-1}$. On the other hand, in the second selection round, the denominator in Eq. \eqref{eqn:solution} is reduced from $(M-2k)$ to $(M-K-2k)$. Therefore, the diversity of selected samples in the second round would be enhanced, according to the following equation.
\begin{equation}
\label{eq3}
    (\frac{2k}{M-2k})^{2}(\mathbf{R}_{1}^{k-1})^{2} \leq (\frac{2k}{M-K-2k})^{2}(\mathbf{R}_{2}^{k-1})^{2}.
\end{equation}

Based on Eq. \eqref{eq3}, the twice selection can be easily extended into recursive selection, so the dataset is divided into several bins with different diversity levels.
The visualizations of recursive selection are shown in the center of Fig.~\ref{fig:ppl}, which also aligns with our analysis well.
We provide more visualizations in the Appendix.

\subsection{Overview of DQ}
Based on the above observation and analysis, we propose Dataset quantization (DQ), a novel framework to quantize large-scale datasets for lossless training, where data efficiency, scalability and computation cost are well considered.
In this paper, we first divide the dataset into several non-overlapping bins by maximizing submodular gains~\cite{iyer2021submodular}.
Specifically, as shown in Fig.~\ref{fig:ppl}, given a  dataset $\mathbf{D}$, small informative bins are sampled from 
$\mathbf{D}$ \textit{recursively} with a pre-defined bin size $K$, yielding a set of small bins $[\mathbf{S}_1, \ldots, \mathbf{S}_n, \ldots, \mathbf{S}_N]$ with $N = M /K$.
Each bin  $\mathbf{S}_n = \{(x_j^{(n)}, y_j^{(n)})\}_{j=1}^K \subset \mathbf{D}$ is constrained under both inter-data diversity and representativeness of the original feature distribution during the recursive selection. 
As analyzed in Sec.~\ref{subsec:extractor}, the bins generated in early steps are mainly constrained by the distance to the remained set, while the latter bins are more constrained by the inter-data diversity. 
To better capture the distribution of the full dataset and balance the influence from the above two perspectives, we then integrate a coreset $\mathbf{S}^{*}$ for training from these bins via uniform sampling.
Eventually, the redundant information is removed by dropping non-informative patches from the images to further reduce the storage burden.

\textbf{Dataset bin generation} Each bin is selected by maximizing the submodular gain \cite{iyer2021submodular} claimed in Eq. \eqref{eqn:graphcut}. DQ recursively selects bins from $\mathbf{D}$, where the selection of $i$-th sample in the $n$-th bin is formulated as follows,

\begin{equation}
    x_{k} \leftarrow \argmax (\sum_{p \in \mathbf{S}_{n}^{k-1}}C_{1}(x_{k}) - \sum_{p \in \mathbf{D} \backslash \mathbf{S}_{1}\cup \cdots  \cup \mathbf{S}_{n}^{k-1}}C_{2}(x_{k})),
\end{equation}
where $C_{1}(x_{k})$ and $C_{2}(x_{k})$ have been defined in Eq. \eqref{eqn:graphcut}, $\mathbf{D} \backslash \mathbf{S}_{1}\cup \cdots  \cup \mathbf{S}_{n}^{k-1}$ denotes the rest of the data in the dataset after selecting
$(k-1)$ samples in $n$-th bin.
We iteratively select the $x$ with the largest submodular gain to form bin $\mathbf{S}_n$,
as detailed  in Algorithm \ref{alg:code}.
The generated bins contain different samples from each other, and each has an emphasis on trade-offs between representativeness and diversity.

\begin{algorithm}[t]
\caption{Data bin generation. }
\label{alg:code}
\begin{algorithmic} %[1]
\STATE \textbf{Input:} original dataset $\mathbf{D}$,   bin number $N$, bin size $K$, the  score function $P(\cdot)$.  
\STATE For $n = 1, \ldots, N-1$ \COMMENT{Indices of sequentially selection}
\STATE \qquad $\mathbf{S}_n^1 \leftarrow \emptyset$, $\mathbf{S}_n^0 \leftarrow \emptyset$   \COMMENT{Initialization of $\mathbf{S}_n$}
\STATE \qquad For $k = 1, \ldots, K$  \COMMENT{Find $K$ most informative samples for $\mathbf{S}_n$}
\STATE \qquad \qquad For $x_{i} \in \mathbf{D} \backslash \mathbf{S}_{n}^{k}$, calculate submodular gains $P(x_{i})$ using Eq. \ref{eqn:graphcut}
\STATE \qquad \qquad \qquad $x^* \leftarrow$ $\argmax_{x \in \mathbf{D} \backslash
\mathcal{S}_{n}^{k}}P(x_{i})$
\STATE \qquad \qquad \qquad $\mathbf{S}_n^k \leftarrow \mathbf{S}_n^{k-1} \cup x^*$
\STATE \textbf{Output:} $N$ dataset bins $\mathbf{S}_1,\ldots\mathbf{S}_n, \ldots \mathbf{S}_N$.
\end{algorithmic}
\end{algorithm}

\textbf{Bin sampling}
After generating the dataset bins with various characteristics, to obtain diverse and informative subset, a sampler $\mathbf{g}(\cdot, \cdot)$ is used to  sample a certain portion from each bin and form the final compact set.
The process is formally defined as:
\begin{equation}
\label{eqn:sampling}
    \mathbf{S^*} = g(\mathbf{S}_1, \rho) \cup \cdots \cup g(\mathbf{S}_n, \rho) \cup \cdots  \cup g(\mathbf{S}_N, \rho),
\end{equation}
where $\rho$ denotes the data keep ratio.
We set $\mathbf{g}(\cdot, \cdot)$ as the uniform sampler by default.

Furthermore, we remove the
redundant data within each sample by dividing them into patches.
Motivated by the Masked Auto-Encoder (MAE) \cite{he2022masked}, which recovers images with only some patches of them, we drop less important patches to reduce the number of pixels utilized for describing each image.
We set $\theta$ as the patch drop ratio and evaluate its sensitiveness in experiments section.
When the data is required for training, the patches are passed through a strong pre-trained MAE decoder to reconstruct the images.
The detailed patch dropping strategy is presented in the Appendix.

\label{method}

\section{Experiments}

\begin{figure*}[t]
\centering
    \begin{subfigure}[]{0.24\textwidth}
    \includegraphics[width=\textwidth, height=0.9\textwidth]{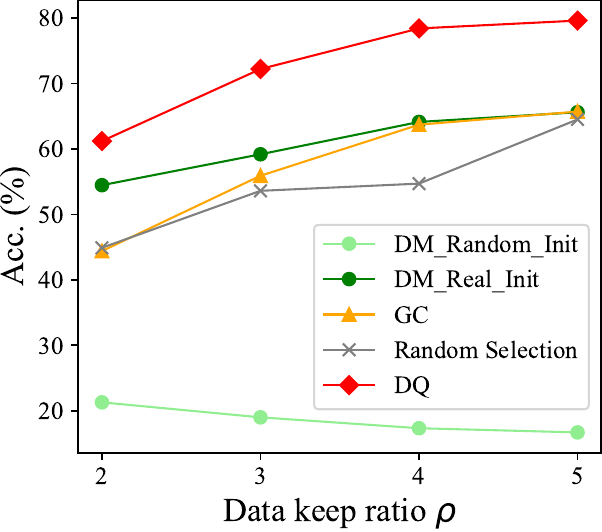}
    \caption{Low Ratio}
    \label{fig:3a}
    \end{subfigure}
    \begin{subfigure}[]{0.24\textwidth}
    \includegraphics[width=\textwidth, height=0.9\textwidth]{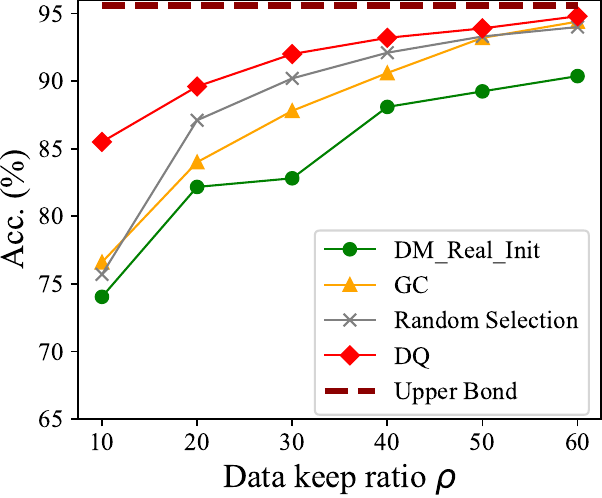}
    \caption{High Ratio}
    \label{fig:3b}
    \end{subfigure}
    \begin{subfigure}[]{0.24\textwidth}
    \includegraphics[width=\textwidth, height=0.9\textwidth]{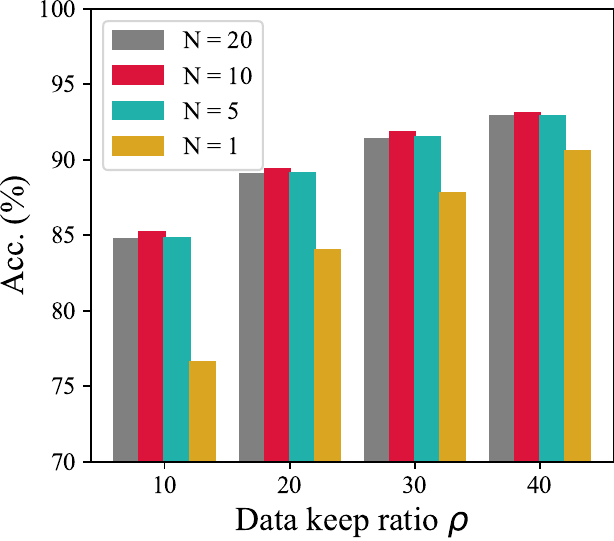}
    \caption{Ablation on $N$}
    \label{fig:3c}
    \end{subfigure}
    \begin{subfigure}[]{0.24\textwidth}
    \includegraphics[width=\textwidth, height=0.9\textwidth]{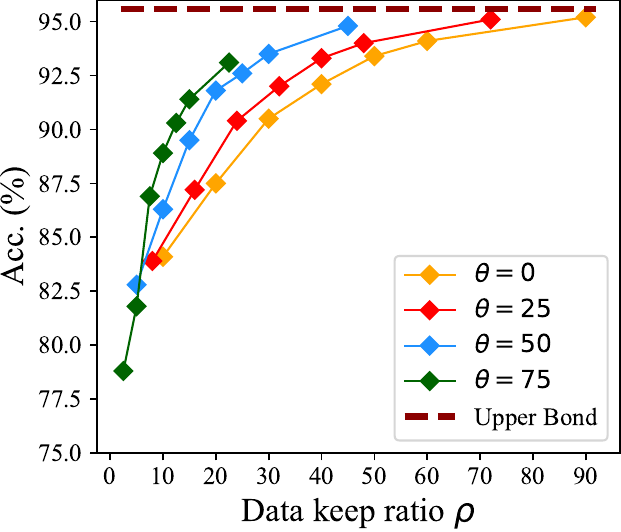}
    \caption{Ablation on $\theta$ (\%)}
    \label{fig:3d}
    \end{subfigure}
    \caption{\textbf{Testing performance of DM~\cite{zhaodm}, random selection, GC~\cite{iyer2021submodular} and DQ on CIFAR-10} at (a) low and (b) high data keep ratio; and   sensitiveness of DQ performance w.r.t.\ (c) the bin number ${N}$  and (d) patch drop ratio $\theta$ across varying data keep ratios. All results are averaged over three runs. The x-axes represent the data keep ratio. 
    \label{fig:plot}} 
\label{fig:data_scability}
\end{figure*} 
\subsection{Datasets and Implementation details}
\textbf{Datasets} We mainly evaluate the proposed dataset quantization method on image classification datasets CIFAR-10 \cite{krizhevsky2009learning} and ImageNet-1K \cite{deng2009imagenet}. CIFAR-10 consists of tiny colored natural images with the size of 32$\times$32 of 10 categories. In CIAFR-10, 50,000 images are used for training and 10,000 images for testing. ImageNet-1K includes 128,1126 images from 1000 categories for training and each category has 50 images for validation. Here, we report the results of the validation set. To better evaluate the transferability  of the pre-trained weights on the compressed dataset from DQ, we also conduct experiments on downstreaming tasks including semantic segmentation and object detection on ADE20K \cite{zhou2019semantic} and COCO \cite{lin2014microsoft}. We report mAP and Seg-mAP on COCO. For segmentation experiments on ADE20K, we report mIoU and aACC.

For large language model (LLM) instruction fine-tuning, we use the alpaca dataset~\cite{alpaca}, which consists of 52k instructions. The alpaca dataset is generated by the self-instruct~\cite{wang2022self} method. To evaluate the fine-tuned LLMs, we follow the benchmark proposed in~\cite{chia2023instructeval}, which consists of MMLU~\cite{hendrycks2020measuring}, BBH~\cite{srivastava2022beyond}, DROP~\cite{dua2019drop}, and HumanEval~\cite{chen2021evaluating} datasets.

\textbf{Implementation details}
\label{sec:implementation_details} Following the previous works \cite{kim2022dataset, zhao2021dataset}, we mainly use ResNet-18 \cite{he2016deep} as the model architecture for the ablation studies, unless specified otherwise. When verifying the generalization capability of the compressed dataset, we use ResNet-18 as the feature extractor during data compression and use the compressed dataset to train representative transformer and CNN architectures, including ViT \cite{dosovitskiy2020image}, Swin transformer \cite{liu2021swin}, ConvNeXt \cite{liu2022convnet} and MobilenetV2 \cite{sandler2018mobilenetv2} models with their official   training recipes.  
For experiments of bin generation, we use ResNet-18 and Vision Transformer (ViT-Base) models to extract features of CIFAR-10 and ImageNet-1K, respectively. The models are pre-trained on the corresponding full dataset with 10 epochs. The number of dataset bins $N$ is set to 10 by default. And the drop ratio $\theta$ is set to 25. We use pytorch-cifar\footnote{https://github.com/kuangliu/pytorch-cifar} and timm library \cite{rw2019timm}  for model training on CIFAR-10 and ImageNet-1K datasets. We train 200 epochs for CIFAR-10 with a batch size of 128 and a cosine-annealed learning rate of 0.1. We train ImageNet in DDP manner with the default scripts of different architectures from timm. For downstream tasks, we follow the popular settings of mmdetection \cite{chen2019mmdetection} and mmsegmentation~\cite{mmseg2020} as used in \cite{zhou2022understanding,zhou2021deepvit}. We choose distribution matching \cite{zhaodm} and graph cut (GC) \cite{iyer2021submodular} as two strong baselines, as well as other well-established dataset compression methods.

For LLM instruction tunning, we follow the training process of alpaca~\cite{alpaca}. We fine-tune the LLaMA-7B~\cite{touvron2023llama} model on the sampled datasets with hyper-parameters introduced in~\cite{zhou2023lima} for a smaller dataset. We use OpenAI's Embedding API~\cite{neelakantan2022text} as the feature extractor during data compression.

\begin{table*}[t]
\caption{Comparisons of cross-architecture generalization of DM and DQ on CIFAR-10. The R18 (first column) is the source architecture used to obtain distilled data or $\mathbf{S^*}$. All architectures are trained from scratch. The top-1 accuracy is reported. CNext stands for the ConvNext architecture. }
\vspace{-1em}
    \centering
    \small
    \begin{subtable}[]{0.455\linewidth}
    \caption{DM on CIFAR-10. }
    \label{tab:5a}
    \resizebox{\linewidth}{!}{
    \begin{tabular}{l| c c c c c c}
    \toprule
         $\rho$ (\%) & R18 & R50 & ViT & Swin & CNext & Avg. \\
         \midrule
        10 & 74.0 & 35.0 & 21.6 & 25.1 & 41.8 & 39.5 \\
        20 & 82.2 & 36.2 & 25.5 & 30.1 & 48.3 & 44.5 \\
        30 & 82.8 & 43.9 & 23.1 & 27.3 & 47.9 & 45 \\
        100 & 95.6 & 95.5 & 80.2 & 90.3 & 73.0 & 86.9 \\
    \bottomrule
    \end{tabular}}
    \end{subtable}
    \begin{subtable}[]{0.515\linewidth}
    \caption{DQ on CIFAR-10. }
    \label{tab:5b}
    \resizebox{\linewidth}{!}{
    \begin{tabular}{l|c c c c c c}
    \toprule
         $\rho$ (\%) & R18 & R50 & ViT & Swin & CNext & Avg. \\
         \midrule
        10 & 84.1 & 82.7 & 58.4 & 69.2 & 52.8 & 69.4 \textcolor{red}{(+29.9)} \\
        20 & 87.6 & 88.1 & 66.8 & 79.1 & 61.8 & 76.7 \textcolor{red}{(+32.2)} \\
        30 & 91.0 & 90.8 & 72.0 & 84.4 & 64.2 & 80.5 \textcolor{red}{(+35.5)} \\
        100 & 95.6 & 95.5 & 80.2 & 90.3 & 73.0 & 86.9 \\
        \bottomrule
    \end{tabular}}
    \end{subtable}
\label{tab:5}
\end{table*}

\begin{table}[t]
    \centering
    \caption{Evaluation of dropping patches 
    randomly and ours with the drop ratio $\theta = 25\%$. \textbf{Bold entries} are best results.}
    \small
    \begin{tabular}{c|cccccc}
    \toprule
         $\rho$ (\%)  & 1 & 3 & 5 &10 & 30 & 50 \\
         \midrule
          Random Acc. (\%) & 41.5  & 69.2 & 77.1 &83.6 & 90.2  & 93.2 \\
           Ours Acc. (\%) & \textbf{42.3}  & \textbf{70.4} &  \textbf{77.8}  & \textbf{84.0} & \textbf{90.6}  & \textbf{93.5} \\
    \bottomrule
     \end{tabular}
    \label{tab:3a}
\end{table}

\begin{table}[t]
    \centering
    \caption{Evaluation of the GPU hours of DM and DQ. 
    We assign 0 to values that are negligible. 
    }
    \small
     \begin{tabular}{c|c | c c c c c c c}
     \toprule
         $\rho$ (\%) &Bin creation  & 10 & 20  &40  & 60 &Total\\
         \midrule
         DM  & 0 & 7  & 14 &29  & 41 &91\\
          DQ &1 & 0 &0& 0  & 0 &1 \\
          \bottomrule
    \end{tabular}
    \label{tab:3b}
\end{table}

\begin{table}[t]
    \centering
    \caption{Impacts of DQ on instruction tuning with LLaMA-7B. 
    }
    \small
     \begin{tabular}{c| c c c c c c c}
     \toprule
         $\rho$ (\%)  & BBH& DROP  &MMLU  & Human-Eval & Avg.\\
         \midrule
         2  &  32.9  & 27.6 & 36.6  & 8.5 & 26.3 \\
          20 & 32.7 & 26.7  & 39.8 & 9.2 & 27.1 \\
        \midrule
         100  &  32.9  & 26.3 & 41.6  & 10.0 & 27.7 \\
          \bottomrule
    \end{tabular}
    \label{tab:3c}
\end{table}

\subsection{Analysis}
\label{subsec:ablations}
In this section, we investigate the effects of different components of DQ and provide apple-to-apple comparisons among DQ, DM \cite{zhaodm} and GC \cite{iyer2021submodular}. 

\textbf{Hyper-parameter analysis.} There are two hyper-parameters  for DQ: the number of bins $\mathbf{N}$ and the drop ratio $\mathbf{\theta}$. We run the experiments with four different values of the bin number: 1, 5, 10, and 20. As shown in Fig. \ref{fig:3c}, the performance drops significantly when the bin number is set to 1. This is the same case of coreset selection where the dataset distribution is not quantized. This gap comes from the fact that a one-time subset selection has limited diversity. When the number of bins is too large, our DQ degrades into random selection, so the performance is worse than our default setting. 
$\theta$ is the patch drop ratio. With a fixed dataset bin number ($N=10$), we vary the drop ratio and the results are shown in Fig. \ref{fig:3d}.  It is observed that a large drop ratio improves the model training performance at large data keep ratio but the performance drops significantly at small data keep ratio. We empirically observe that the combination of $N=10$ and $\theta=25\%$ give   the best trade-off.

\textbf{Generalizability of the compressed datasets.}
We  investigate the generalizability of the compressed datasets for training different architectures. 
Fig. \ref{fig:1c} has demonstrated DQ can well preserve the dataset distribution for various architectures. 
We further look into the impact on the quantitative performance.  We use DQ and DM to compress the dataset by 90\%, 80\% and 70\% respectively, and use the generated dataset to train the selected models as detailed in Sec. \ref{sec:implementation_details}. The results are shown in Tab.~\ref{tab:5}. As observed, under all data keep ratios, the dataset generated by DM suffers a significant performance drop when trained on unseen architectures. 
The drop is relatively small on CNN models and larger on transformer-based models. When used for training the ViT and Swin models, the performance drops by up to 70 percentage with DM generated dataset. In contrast, the datasets compressed by DQ offer better performance. 
Tab. \ref{tab:5b} shows the average benefits of DQ relative to DM in the final column. In average, DQ performs better than DM by a range of 29.9\% to 35.5\% under different data keep ratios.
It validates that the compression process of DQ is model-agnostic, indicating better generalizability.

\textbf{Compression scalability.} We investigate how  the performance of different compression methods changes under different data keep ratio.
We use DQ, DM and GC to compress the CIFAR-10 dataset to the same ratio and then use the compressed dataset to train ResNet-18 from scratch. The results under low and high ratios are shown in Fig. \ref{fig:3a} and \ref{fig:3b}, respectively.
It is clearly observed that when the data keep ratio ratio is extremely low (e.g. 1\%), the coreset based algorithm GC gives the lowest accuracy. Under high data keep ratio, the dataset distillation-based method DM saturates quickly and the final accuracy is 5\% lower than the random sampling. Under both cases, DQ achieves the highest accuracy when used for model training, demonstrating outperforming scalability.

\textbf{Impact of the image patch attention.} 
As mentioned above, we calculate a patch importance score to drop less important patches to decrease the redundancy of the original dataset. Randomly removing these patches is a simple and basic approach. We compare the efficacy of randomly dropping patches versus using GradCAM-based drops. As illustrated in Tab. \ref{tab:3a}, our method outperforms the random strategy for all data retention ratios.
More details are provided in the Appendix. 
\begin{figure*}[t]
    \centering
    \begin{subfigure}[]{0.48\textwidth}
    \tiny
    \begin{overpic}[width=\textwidth]{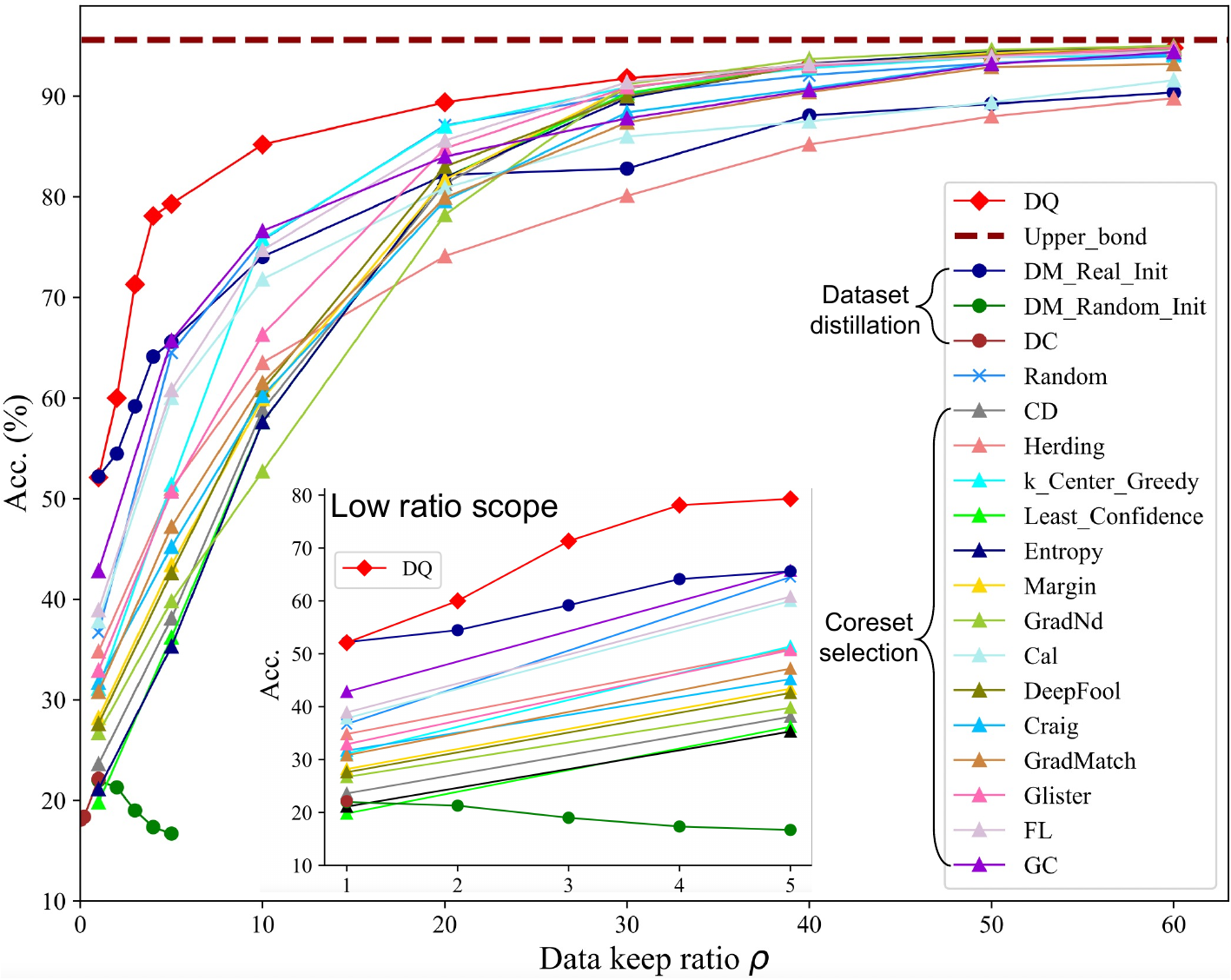}
    \end{overpic}
    \caption{}
    \label{fig:cross_compa}
    \end{subfigure}
    \begin{subfigure}[]{0.48\textwidth}
    \tiny
    \begin{overpic}[width=\textwidth]{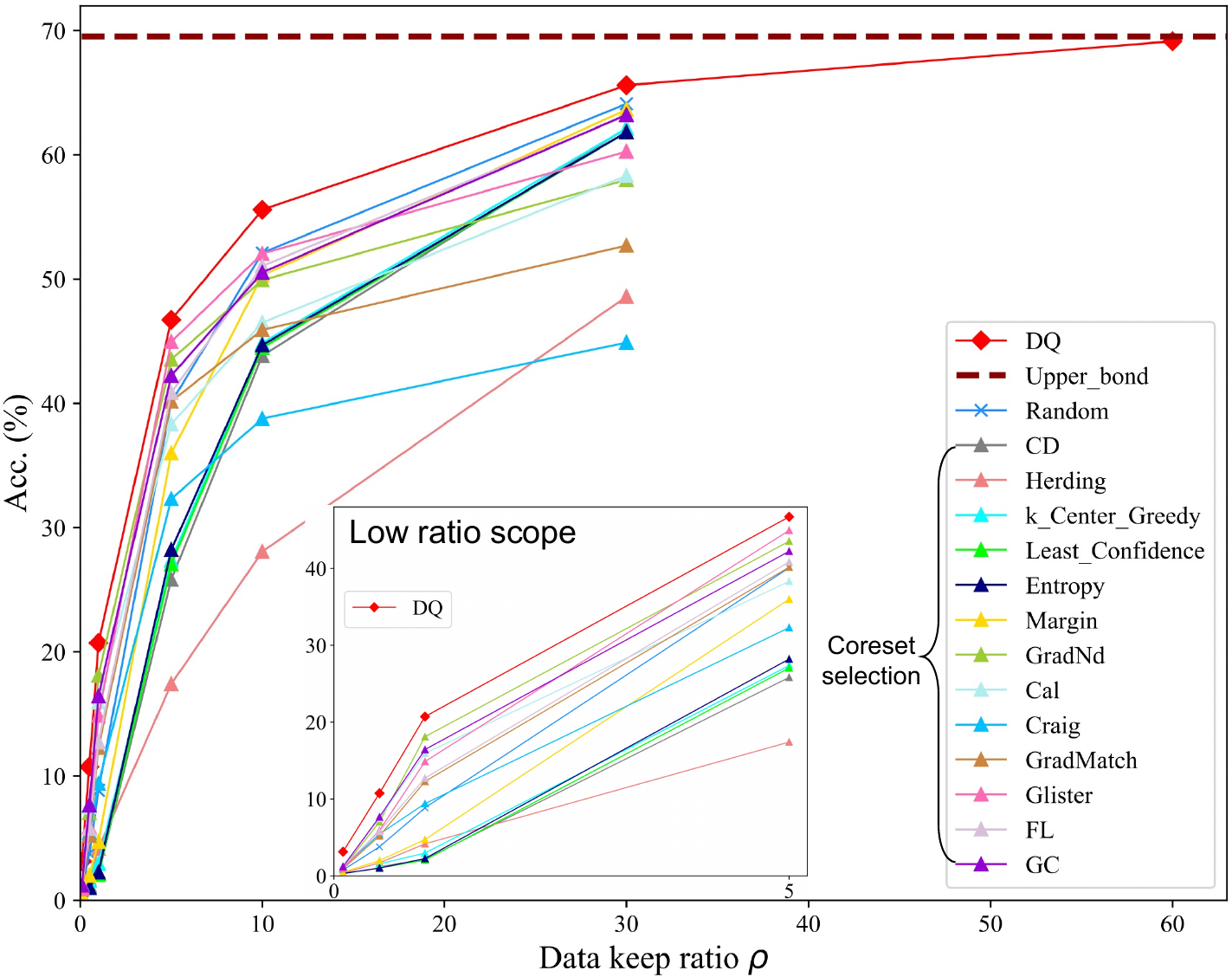}
    \end{overpic}
    \caption{}
    \label{fig:cross_compb}
    \end{subfigure}
\vspace{-1em}
\caption{Comparison of DQ with previous state-of-the-arts with different data keep ratios on (a) CIFAR-10 and (b) ImageNet-1K. \label{fig:cross_comp}
}
\end{figure*}

\begin{figure*}[t]
    \centering
    \begin{subfigure}[]{0.3\textwidth}
    \tiny
    \begin{overpic}[width=\textwidth, height=0.8\textwidth]{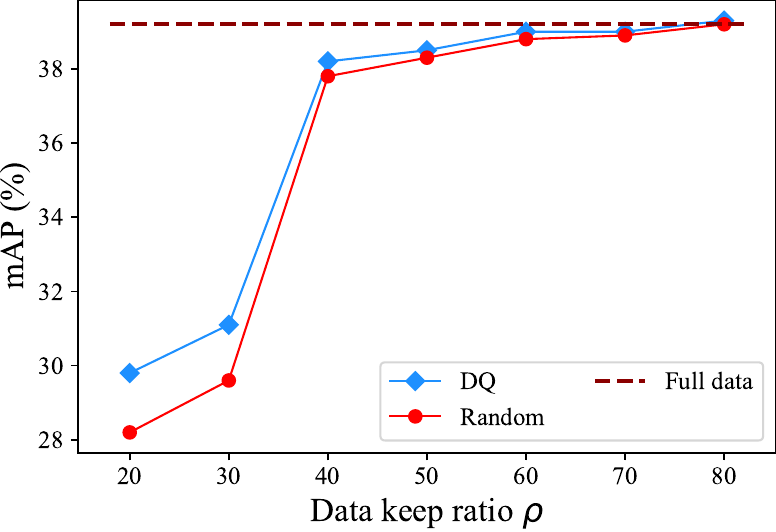}
    \put(48,45){
    \tiny
    \setlength\tabcolsep{0.5mm}
    \renewcommand{\arraystretch}{1}
    \begin{tabular}{c|c} 
    \toprule
      $\rho$(\%)   & Seg\_mAP (\%)  \\ \hline
      20 &24.4\\
      30 &28.5\\
      40  &34.7\\
      50  &34.8\\
      60 &35.2 \\
      70 & 35.2\\
      80 & \textbf{35.4} \\
      Full data & \textbf{35.4} \\
      \bottomrule
    \end{tabular}}
    \end{overpic}
    \caption{}
    \end{subfigure}
    \begin{subfigure}[]{0.3\textwidth}
    \tiny
    \begin{overpic}[width=\textwidth, height=0.8\textwidth]{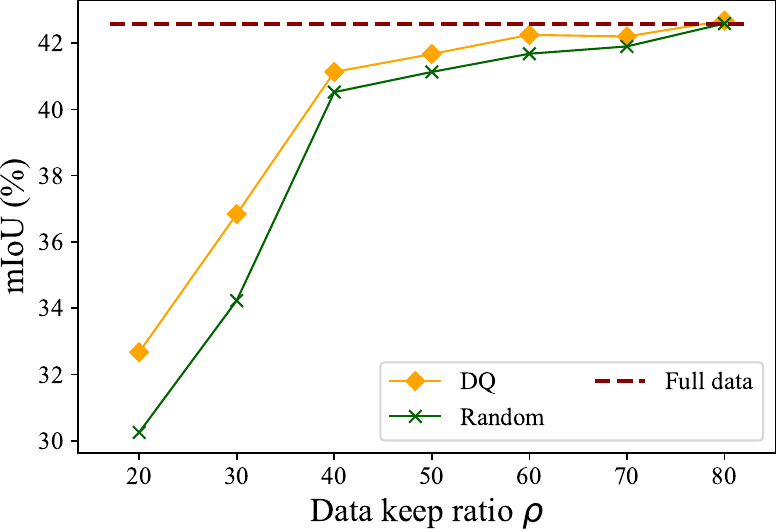}
    \put(48,45){
    \tiny
    \setlength\tabcolsep{0.5mm}
    \renewcommand{\arraystretch}{1}
    \begin{tabular}{c|c} 
    \toprule
      $\rho$(\%)   & aAcc (\%)  \\ \hline
      20 &73.9\\
      30 &78.1\\
      40  &80.0\\
      50  &79.9\\
      60 &80.3 \\
      70 & 80.3\\
      80 & \textbf{80.4} \\
      Full data & 80.3 \\
      \bottomrule
    \end{tabular}}
    \end{overpic}
    \caption{}
    \end{subfigure}
    \begin{subfigure}[]{0.3\textwidth}
    \begin{overpic}[width=\textwidth, height=0.8\textwidth]{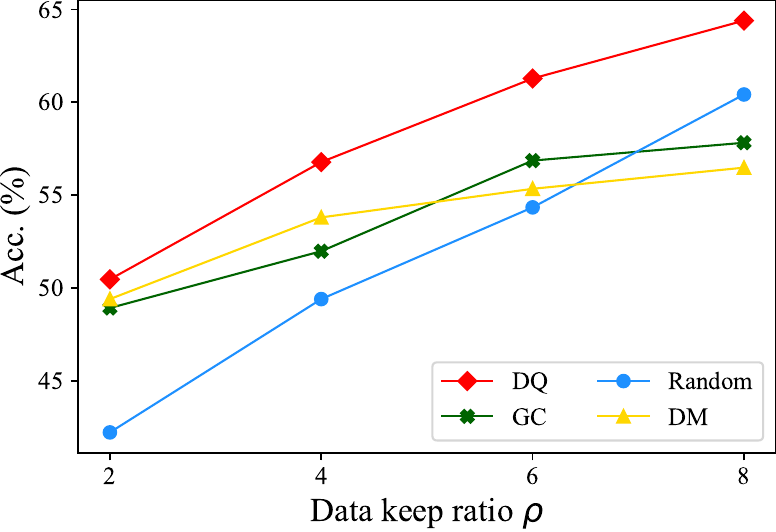}
    \end{overpic}
    \caption{}
    \label{fig:rbst}
    \end{subfigure}
\vspace{-1em}
\caption{Comparisons of the performances for  downstream tasks on (a) COCO and (b) ADE20K. (c) Comparisons of the robustness of trained models via random selection, DM~\cite{zhaodm}, GC~\cite{iyer2021submodular}, and DQ on CIFAR10-C dataset. We report average performances of 15 types of corruption on 5 levels. More detailed results can be found in Appendix.
}
\label{fig:downstream}
\vspace{-2mm}
\end{figure*}

\textbf{Computational cost analysis.} Due to the synthesizing strategy used in DM, a large tensor (\textit{i.e.} initialization of synthetic images) needs to be defined. As a result, both the computational cost and memory consumption increase linearly  to the size of the dataset. We directly measure the GPU hours needed for synthesizing the dataset and the results are shown in Tab. \ref{tab:3a}. 
Once the data keep ratio changes, the whole process of DM needs to be repeated.
In contrast, DQ only needs to quantize the whole dataset into several bins. 
The following sampling step takes negligible GPU computations (N.A. in Tab. ~\ref{tab:3a}).

\subsection{Comparison with state-of-the-art methods}
We compare our method to previous state-of-the-art methods on both CIFAR-10 and ImageNet. 
We compare our proposed DQ with 3 dataset distillation and 14 coreset selection methods.
The results are shown in Fig. \ref{fig:cross_comp}. We would like to highlight that the results of dataset distillation methods are only shown on CIFAR-10 dataset.
Due to the extremely large computational cost, it is not feasible to verify those methods on the ImageNet-1K dataset intuitively.
The computational cost of dataset distillation methods can be checked from Fig. \ref{fig:1a}.
To better understand the characteristics of DQ, we also scope the performance comparisons under low data keep ratios.
DQ outperforms other coreset selection and dataset distillation methods by a large margin, which indicates 
that DQ provides stronger data efficiency under the same data keep ratio. 
Our method is based on GC algorithm, while outperforms GC by a large margin on all data keep ratios. On both CIFAR-10 and ImageNet-1K, we obtain lossless results when using only 60\% data, setting a new state-of-art for dataset compression.
Actually, DQ works as a play-and-plug module that could be combined with most coreset selection methods.

\subsection{Performance on language tasks}
To evaluate the effectiveness of DQ on language tasks, we choose four popular benchmarks of BBH, DROP, MMLU, and Human-Eval, following Alpaca. The results are shown in Tab. \ref{tab:3c}. As observed, with only 20\% of the instruction tuning data extracted with DQ, comparable performance can be achieved as the model is finetuned with full data.

\subsection{Performance on downstream tasks}
To further evaluate the data efficiency of DQ on downstream tasks, we finetune the pretrained models with different data keep ratios (from 20\% to 80\%) on COCO and ADE20K datasets. 
Here, we make a comparison with random selection strategy.
As shown in Fig. \ref{fig:downstream}, our proposed DQ   achieves comparable mAP and mIOU results as training on full data when the data keep ratio is 60\%.
Setting the data keep ratio as 80\% can achieve lossless results, which indicates the samples selected by DQ are informative. 
From 20\% to 40\% data keep ratio, DQ achieve obvious higher results than random selection strategy. 
We would like to highlight that this is not feasible for DM or other dataset distillation methods due to the unaffordable computation cost to compress ImageNet and obtain the pre-trained model  as mentioned in Sec. \ref{subsec:ablations}.

\subsection{Robustness Evaluation} 
In order to investigate the robustness of our proposed DQ, we compared its performance with that of GC and Random selection by evaluating their performance on the corrupted dataset of CIFAR10 with the same methods used in  ~\cite{hendrycks2019robustness,zhou2022understanding}. The results, depicted in Figure \ref{fig:rbst}, demonstrate that our proposed DQ method achieves superior results at all data retention ratios. The performance gap between GC and random selection narrows as the data retention ratio increases, which can be attributed to the fact that the samples selected by GC lack diversity. More detailed results can be found in Appendix.

\section{Conclusion}
We present a new dataset compression pipeline, termed dataset quantization (DQ), that is able to achieve lossless compression and be used to train unseen network architectures. We conduct extensive experiments showing that DQ achieves new state-of-the-art compression ratios. For the first time, we verify that models pre-trained with the compressed dataset can be used for training downstream tasks such as object detection and semantic segmentation. We hope this work could motivate more research works toward more generalizable dataset compression algorithms.

\textbf{Limitations and future works}
Our DQ needs to select samples recursively from the whole dataset, resulting in extra computational efforts. In the future, we aim to design a more advanced DQ that only selects once from the whole dataset. Meanwhile, we plan to explore DQ on other tasks, such as video understanding~\cite{soomro2012ucf101}, AIGC~\cite{bain2021frozen}, and so on.

\section*{Acknowledgement}
This research is supported by the National Research Foundation, Singapore under its AI Singapore Programme (AISG Award No: AISG2-PhD-2021-08-
008). Yang You's research group is being sponsored by NUS startup grant (Presidential Young Professorship), Singapore MOE Tier-1 grant, ByteDance grant, ARCTIC grant, SMI grant and Alibaba grant.

\textit{Special Acknowledgement}. We would like to thank \textbf{Zangwei Zheng} for his help on the implementation of DQ in language tasks and \textbf{Ge Yan} for his advice on the mathematical proof of the submodular part. 
\label{exp}

{\small
\bibliographystyle{ieee_fullname}
\bibliography{iclr2023_conference}
}

\onecolumn
\section{Appendix}
We present more explanations of the proposed \nameofmethod, experiment results and visualizations in this section. 

\subsection{Proof of Sec. 3.1}
\label{appendix:proof}
Given the whole dataset $\mathbf{D}$, $|\mathbf{D}| = M \gg 1$. $\forall{p} \in \mathbf{D}$, $f(p) \in \mathbb{R}^{m \times 1}$.
To make a simple proof, we assume $\frac{1}{M}\sum_{p \in \mathbf{D}}f(p) = 0$.

For $n = 0, 1, 2, \cdots, n, \cdots, N$, define set $\mathbf{S}_{n} \in \mathbf{D}$. And, we define $C_{1}(x)$ and $C_{2}(x)$ as follows,

\begin{equation}
    C_{1}(x) = \sum_{p \in \mathbf{S}_{1}^{n}}||f(p) -f(x)||_{2}^{2}; \quad \quad \quad C_{2}(x) = \sum_{p \in \mathbf{D} \backslash \mathbf{S}_{1}^{i}}||f(p) -f(x)||_{2}^{2}; 
\end{equation}

By the policy of GraphCut~\cite{iyer2021submodular}, it aims to maximize $C_{1}(x)$ and minimize $C_{2}(x)$ to select $\mathbf{S}_{1}$. We write it into a united target function to choose $x_{k+1}$ as,

\begin{equation}
    x_{k+1} \leftarrow \argmax_{x \in \mathbf{D} \backslash \mathbf{S}_{1}^{k}}(C_{1}(x) - C_{2}(x)).
\end{equation}
We initialize $\mathbf{S}_{1}^{1}$ using $\emptyset$ and $\mathbf{S}_{1}^{k+1} = \mathbf{S}_{1}^{k} \cup {x_{k+1}}$, where $k = 1, 2, \cdots, k, \cdots, K$, and $|\mathbf{S}_{1}^{k}| = k$.

\textbf{Claim:} (a). $\mathbf{S}_{1}^{1} = \emptyset$, $x_{1} = \argmin_{x \in \mathbf{D}}||x||_{2}^{2}$, \textit{i.e,} the closest point to \textbf{0} in $\mathbf{D}$

(b). $x_{k+1}$ is very close to set $\mathbf{S}_{1}^{k}$.

\textbf{Proof:} $\mathbf{S}_{1}^{1} = \emptyset$, so $C_{1}(x) = 0$.
\begin{align}
    C_{2}(x) &= \sum_{p \in \mathbf{D}}||f(p)-f(x)||_{2}^{2}\\ 
    &= M||f(x)||_{2}^{2} - 2(\sum_{p \in \mathbf{D}}f(p))^{\top}f(x) + \sum_{p \in \mathbf{D}}||f(p)||_{2}^{2} \\
&= M||f(x)||_{2}^{2} + \sum_{p \in \mathbf{D}}||f(p)||_{2}^{2},
\end{align}
where $\sum_{p \in \mathbf{D}}f(p) = 0$. 

Then, we have 
\begin{align}
  x_{1} &= \argmax_{x \in \mathbf{D}} -C_{2}(x)\\
   &= \argmin_{x \in \mathbf{D}}C_{2}(x)\\
   &= \argmin_{x \in \mathbf{D}}M||f(x)||_{2}^{2},
\end{align}

(b) Let $C_{1}(x_{k}) - C_{2}(x_{k}) = 2C_{1}(x_{k}) - (C_{2}(x_{k}) + C_{1}(x_{k}))$.
We have:
\begin{align}
    (C_{2}(x_{k}) + C_{1}(x_{k})) &=\sum_{p \in \mathbf{D} \backslash \mathbf{S}_{1}^{k}}||f(p) - f(x_{k})||_{2}^{2} + \sum_{p \in  \mathbf{S}_{1}^{k}}||f(p) - f(x_{k})||_{2}^{2} \\
    &= \sum_{p \in \mathbf{D}}||f(p)-f(x)||_{2}^{2} \\
    &= M||f(x)||_{2}^{2} + \sum_{p \in \mathbf{D}}||f(p)||_{2}^{2} \\
    &= M||f(x)||_{2}^{2} + Const.,
\end{align}
where `Const.' denotes constant number.

For $C_{1}(x_{k})$, we have
\begin{equation}
   C_{1}(x_{k}) = \sum_{p \in \mathbf{S}_{1}^{k}}||f(p) - f(x_{k})||_{2}^{2} = 
k||f(x_{k})||_{2}^{2} -2(\sum_{p \in \mathbf{S}_{1}^{k}}f(p)^{\top}f(x_{k}) + \sum_{p \in \mathbf{S_{1}^{k}}}||f(p)||_{2}^{2} 
\end{equation}
Define $Q_{k} = \frac{1}{k}\sum_{p \in \mathbf{S}_{1}^{k}}f(p)$ as the weighted center of $\mathbf{S}_{1}^{k}$.
Then, we can write the submodular gains function as follows,

\begin{align}
    P(x_{k}) &= 2C_{1}(x_{k}) - (C_{2}(x_{k}) + C_{1}(x_{k})) \\
    &=2k||f(x_{k})||_{2}^{2} - 4kQ_{k}^{\top}f(x_{k}) - M||f(x_{k})||_{2}^{2} + Const. \\
    &= (2k - M) ||f(x_{k}) - \frac{2kQ_{k}}{2k - 
    M}||_{2}^{2} + Const.
\end{align}

$x_{k+1}$ is selected as follows,

\begin{equation}
    x_{k+1} = \argmax_{x \in \mathbf{D} \backslash \mathbf{S}_{1}^{k}}P(x_{k}) = \argmax_{x \in \mathbf{D} \backslash \mathbf{S}_{1}^{k}}||f(x_{k}) - \frac{2kC_{k}}{2k-M}||_{2}^{2}.
\label{star_eqn}
\end{equation}

Let $\delta_{k} = \frac{2kC_{k}}{2k-M}$. We define radius $R_{1}^{k}$ of set $\mathbf{S_{1}^{k}}$ as,

\begin{equation}
    R_{1}^{k} = max_{p \in \mathbf{S}_{1}^{k}}||f(p)||_{2}.
\end{equation}
Therefore, $\forall{p} \in \mathbf{S}_{1}^{k}$,
$||f(p)||_{2}^{2} \leq (R_{1}^{k})^2$, which means $\mathbf{S}_{1}^{k}$ is included in a ball $\mathbf{B}_{k} = \{p|||f(p)||_{2}^{2} \leq (R_{1}^{k})^2\}$.
% $\mathbf{B}_{k} = {p|||y||_{2}^{2} \leq R_{k}^2}$.
Note that,
\begin{align}
   ||\delta_k||_{2}^{2} &= (2k/2k-M)^2||Q_{k}||_{2}^{2} \\
   &= (\frac{2k}{2k-M})^{2}||\frac{1}{k}\sum_{p \in \mathbf{S}_{1}^{k}}f(p)||_{2}^{2}\\
   &\leq (\frac{2k}{2k-M})^{2}\frac{1}{k}\sum_{p \in \mathbf{S}_{1}^{k}}||f(p)||_{2}^{2}\\
   &\leq (\frac{2k}{2k-M})^{2}(R_{1}^{k})^{2}.
\end{align}

$M \gg k$, so $||\delta_{k}||_{2}^{2} \leq (R_{1}^{k})^{2}$ and $\delta_{k} \in \mathbf{B}_{k}$
According to Eq. \ref{star_eqn}, $x_{k+1} = \argmin_{x \in \mathbf{D} \backslash \mathbf{S}_{1}^{k}}||x - \delta_{k}||_{2}^{2}$ is the closest point in $\mathbf{D} \backslash \mathbf{S}_{1}^{k}$ to $\delta_{k}$, which is in the ball $\mathbf{B}_{k}$. As $M \gg 1$, $f(x_{k+1})$ is very close to $\mathbf{B}_{k}$, and thus to $\mathbf{S}_{1}^{k}$.

By the proof, GraphCut cannot guarantee the samples diversity under small data keep ratio.
Our DQ recursively select samples from $\mathbf{D}$, as the total number of $\mathbf{D}$ reduces, the radius of the ball $\mathbf{B_k}$ will be extended. Therefore the sample diversity is higher than GraphCut method.

\subsection{Details of Patch Dropping and Reconstruction}
\label{details_patch_attention}
As pointed out in Masked Auto-Encoder (MAE) ~\cite{he2022masked}, with a pre-trained decoder, some image patches can be dropped without affecting the reconstruction quality of the image. Motivated by it, we propose to reduce the number of pixels utilized for describing each image.
Specifically, as shown in pipeline, given an image $x$,
we first feed it into a pretrained feature extractor (ResNet-18 ~\cite{he2016deep}) 
to obtain the last feature map $\mathcal{M}$ and a prediction score $y^c$ of the image class $c$. 
A group of attention scores is then calculated with the gradient values of each pixel in the last feature map following GradCAM++ ~\cite{aditya1710grad}:

\begin{equation}
 a^c = \sum_{i,j}\left[\frac{\frac{\partial ^2y^c}{(\partial \mathcal{M}_{ij})^2}}{2 \frac{\partial ^2y^c}{(\partial \mathcal{M}_{ij})^2}\! + \! \sum_{m,n} \mathcal{M}_{mn}\{\frac{\partial ^3y^c}{(\partial \mathcal{M}_{ij})^3}\}}\right] \mathrm{ReLU}\left(\frac{\partial y^c}{\partial \mathcal{M}_{ij}}\right),
\end{equation}
where $a^c$ is the attention scores for each pixel w.r.t.\ class $c$, $\mathrm{ReLU}$ is the Rectified Linear Unit activation function, and ($i$, $j$) and ($m$, $n$) are iterators over the feature map $A$.
The pixel-wise attention score $a^c$ is upsampled to fully cover the original input image.
In order to integrate the attention information 
into image patches, we unify the attention scores of the corresponding pixels of a patch by their average value to generate the patch-wise importance scores $p^c_{\cdot}$ as follows,
\begin{equation}
p^c_k=\frac{1}{hw}\sum_{i=h_k}^{h_k+h}\sum_{j=w_k}^{w_k+w}a^c(i,j),
\end{equation}
where $h_k$ and $w_k$ are the coordinates of the upper left corner of the patch $k$, and $h$ and $w$ are the height and width of image patches. 
According to the patch-wise attention scores, we drop a percentage of $\theta$ non-informative patches with smallest attention scores to further save the storage cost.
At the training stage, we employ a strong pre-trained MAE decoder to reconstruct the dropped patches and the original images.

\subsection{Robustness Evaluation}
\begin{figure*}[t]
\centering
    \begin{subfigure}{0.19\textwidth}
    \includegraphics[width=\textwidth,height=0.9\textwidth]{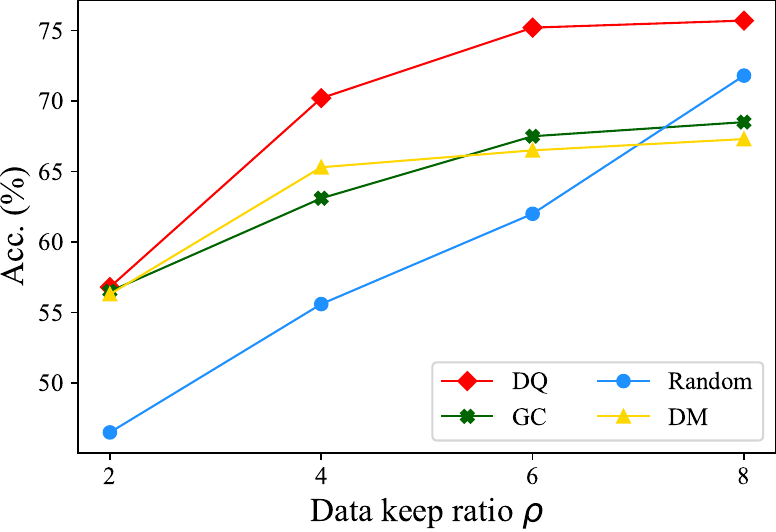}
    \caption{Level 1}
    \label{fig:5a}
    \end{subfigure}
    \begin{subfigure}{0.19\textwidth}
    \includegraphics[width=\textwidth,height=0.9\textwidth]{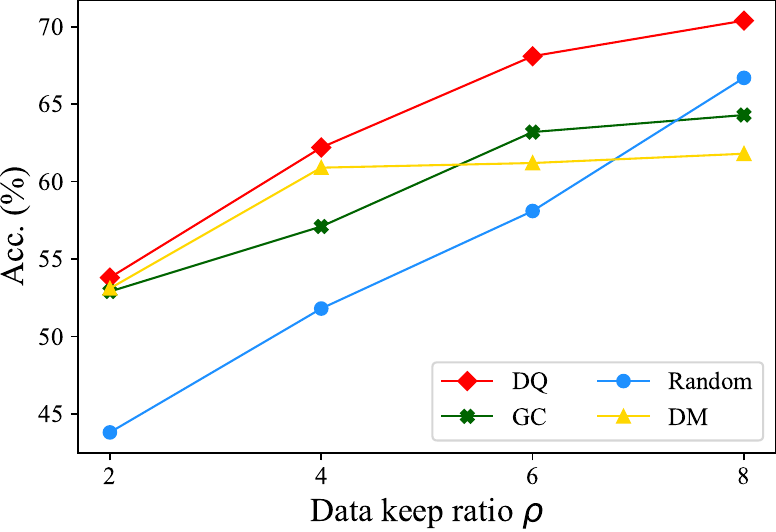}
    \caption{Level 2}
    \label{fig:5b}
    \end{subfigure}
    \begin{subfigure}{0.19\textwidth}
    \includegraphics[width=\textwidth,height=0.9\textwidth]{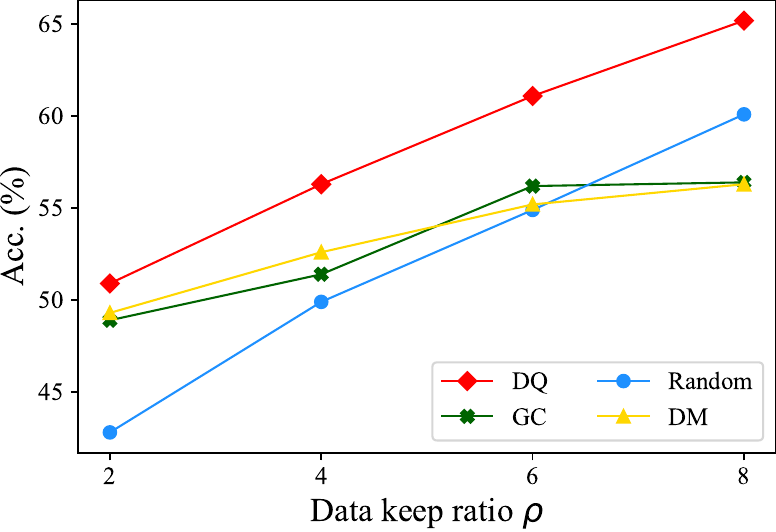}
    \caption{Level 3}
    \label{fig:5c}
    \end{subfigure}
    \begin{subfigure}{0.19\textwidth}
    \includegraphics[width=\textwidth,height=0.9\textwidth]{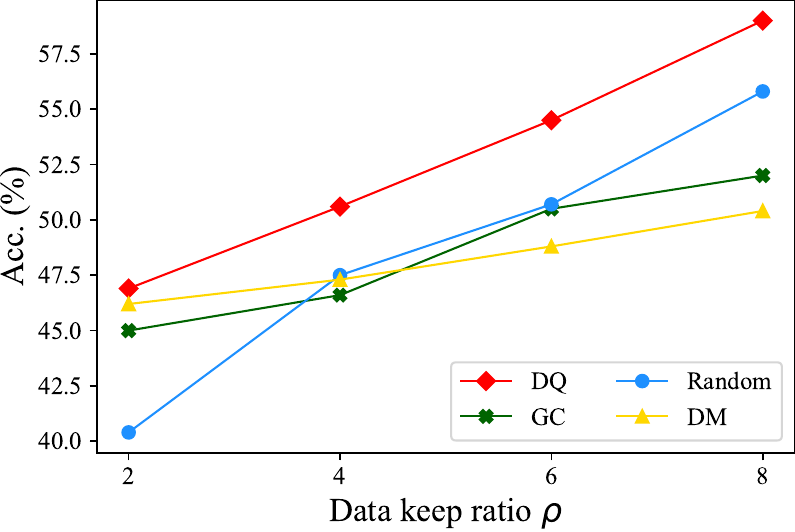}
    \caption{Level 4}
    \label{fig:5d}
    \end{subfigure}
    \begin{subfigure}{0.19\textwidth}
    \includegraphics[width=\textwidth,height=0.9\textwidth]{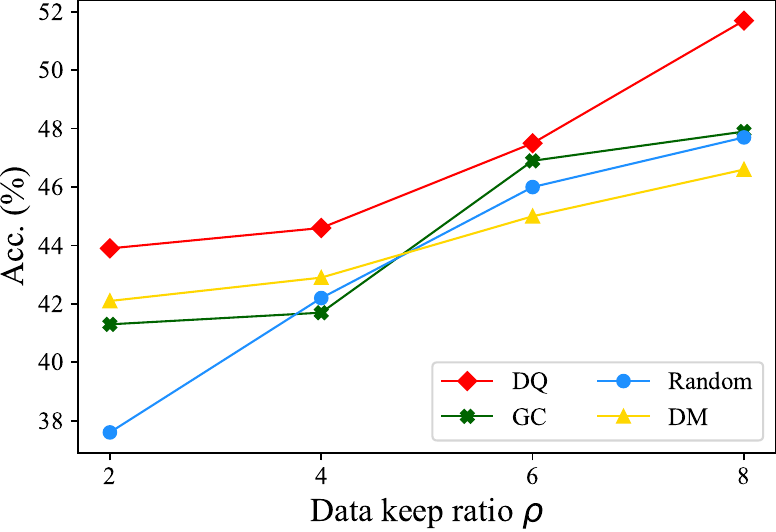}
    \caption{Level 5}
    \label{fig:5e}
    \end{subfigure}
    \caption{Comparisons of the robustness of trained models via DQ, GC and Random selection on CIFAR10-C dataset.} 
\label{fig:fig_robust}
\end{figure*}

We show the overall robustness evaluation in our paper. Here, we report the detailed results at different corruption levels in Fig.~\ref{fig:fig_robust}.
Our proposed DQ achieves state-of-the-art results in all cases.
\label{rbst_appendix}

\subsection{Differences between coreset selection and \nameofmethod}

\begin{figure}[h]
    \centering
    \includegraphics[width=\textwidth]{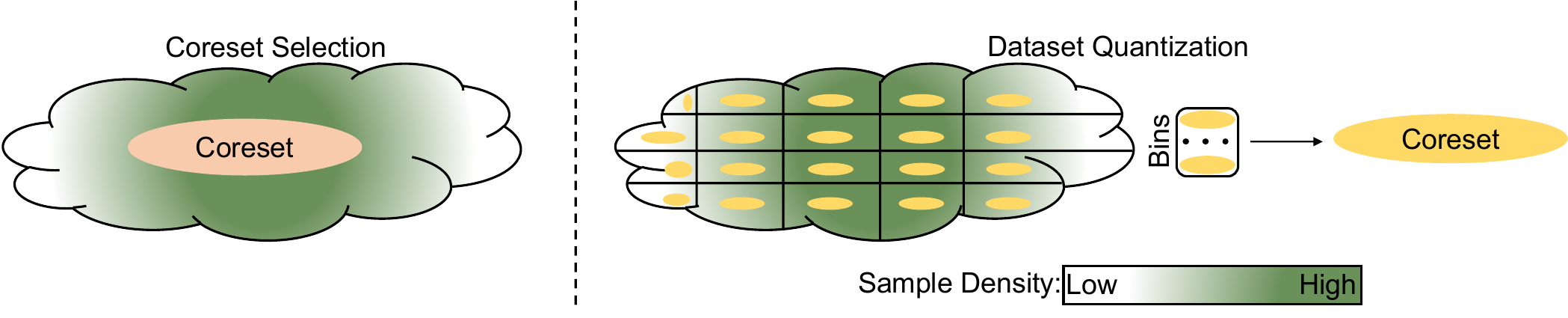}
    \caption{Differences between coreset selection methods and our dataset quantization.}
    \label{fig:coreset_comp}
\end{figure}

\begin{figure}[t]
    \centering
    \includegraphics[width=\textwidth]{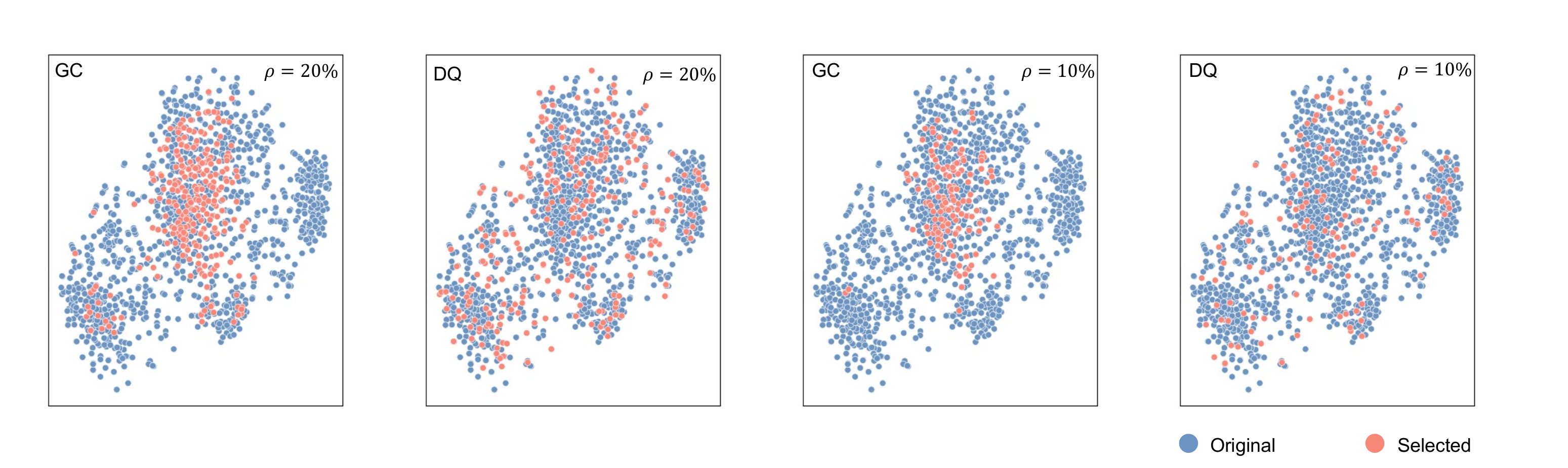}
    \caption{Visualization of the feature distributions among data selected by GraphCut and SQ.}
    \label{fig:vis_gc_dq}
\end{figure}

\paragraph{Coreset VS DQ} We here give more detailed explanations on the difference between the coreset selection methods and our proposed \nameofmethod. As shown in Fig. \ref{fig:coreset_comp}, the coreset selection only select one subset from the full data distribution. This practice will suffer from a selection bias, resulting in selection results with limited diversity. Besides, when the the size of the selected subset is small, it will suffer a large selection variance. Differently, \nameofmethod~first divides the full distribution into non-overlapping bins and then sampling from each bin uniformaly. As a result, the sampled data could maximally preserve the original data distribution. To verify this, we use GraphCut ~\cite{iyer2021submodular} as a representation of the coreset based method and 10\% and 20\% data from 
ImageNet dataset and compare the results with the data distribution sampled with \nameofmethod. We use a pre-trained ResNet-18 model to extract the features of the data and then visualize the extracted data via t-SNE. The results are shown in Fig. \ref{fig:vis_gc_dq}. It is clearly observed that the data sampled via \nameofmethod~ do capture a more diverse distribution.

% \clearpage

\begin{figure}[t]
    \centering
    \includegraphics[width=.95\textwidth]{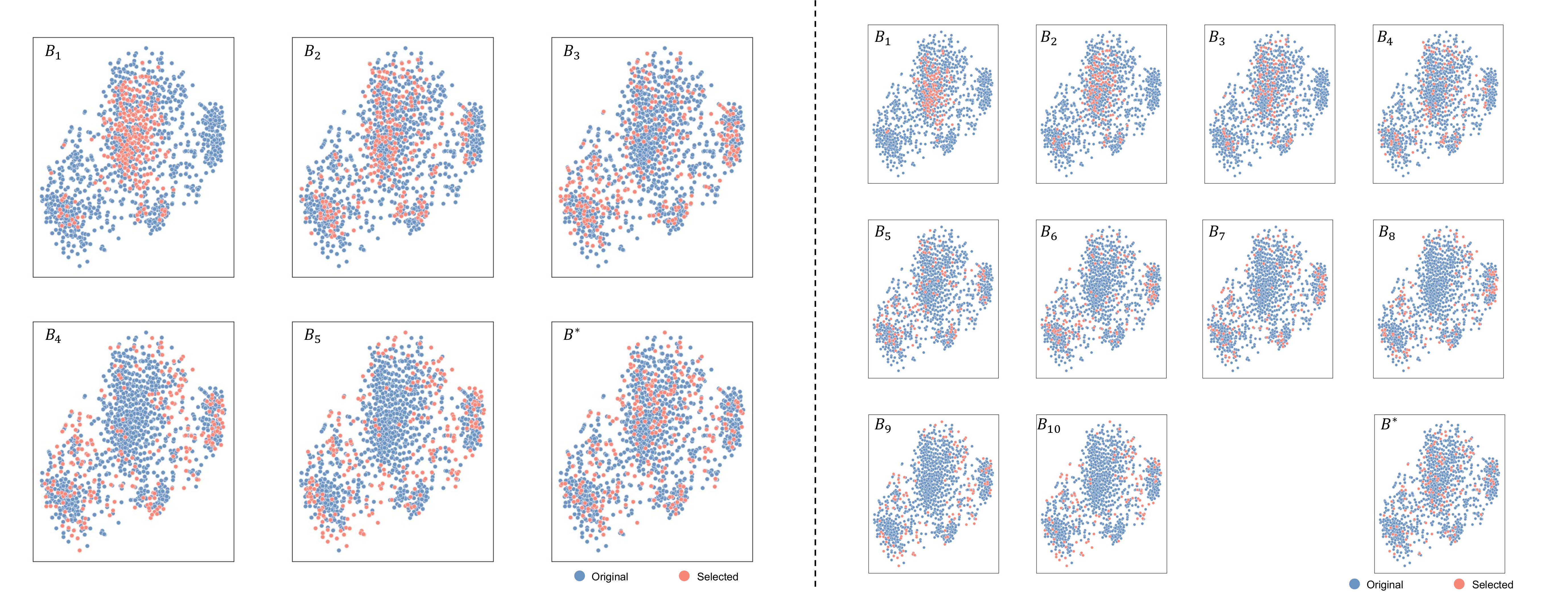}
    \caption{Visualization of the feature distributions among data selected in each bin and the final output of SQ on ImageNet dataset tench class. The bin number $N$ and the data keep ratio $\rho$ are set as (5, 20), (10, 10), respectively for the left and right column. }
    \label{fig:in_010_020_tsne}
\end{figure}

\paragraph{Bin diversity of DQ} To dig deeper for the reason why DQ can better preserved the data distribution. We use the same visualization method as aforementioned for the data contained within each bin. The results are shown in Fig. \ref{fig:in_010_020_tsne}. Each bin contains 20\% of the total data in the left column and 10\% data in the right column. As shown, different bins are capturing different distributions. As a results, after sampling uniformly from each bin, the combined dataset enjoys a large diversity as well as representativeness over the whole data distribution. 

\clearpage

\begin{figure}[ht]
    \centering
    \includegraphics[width=\textwidth]{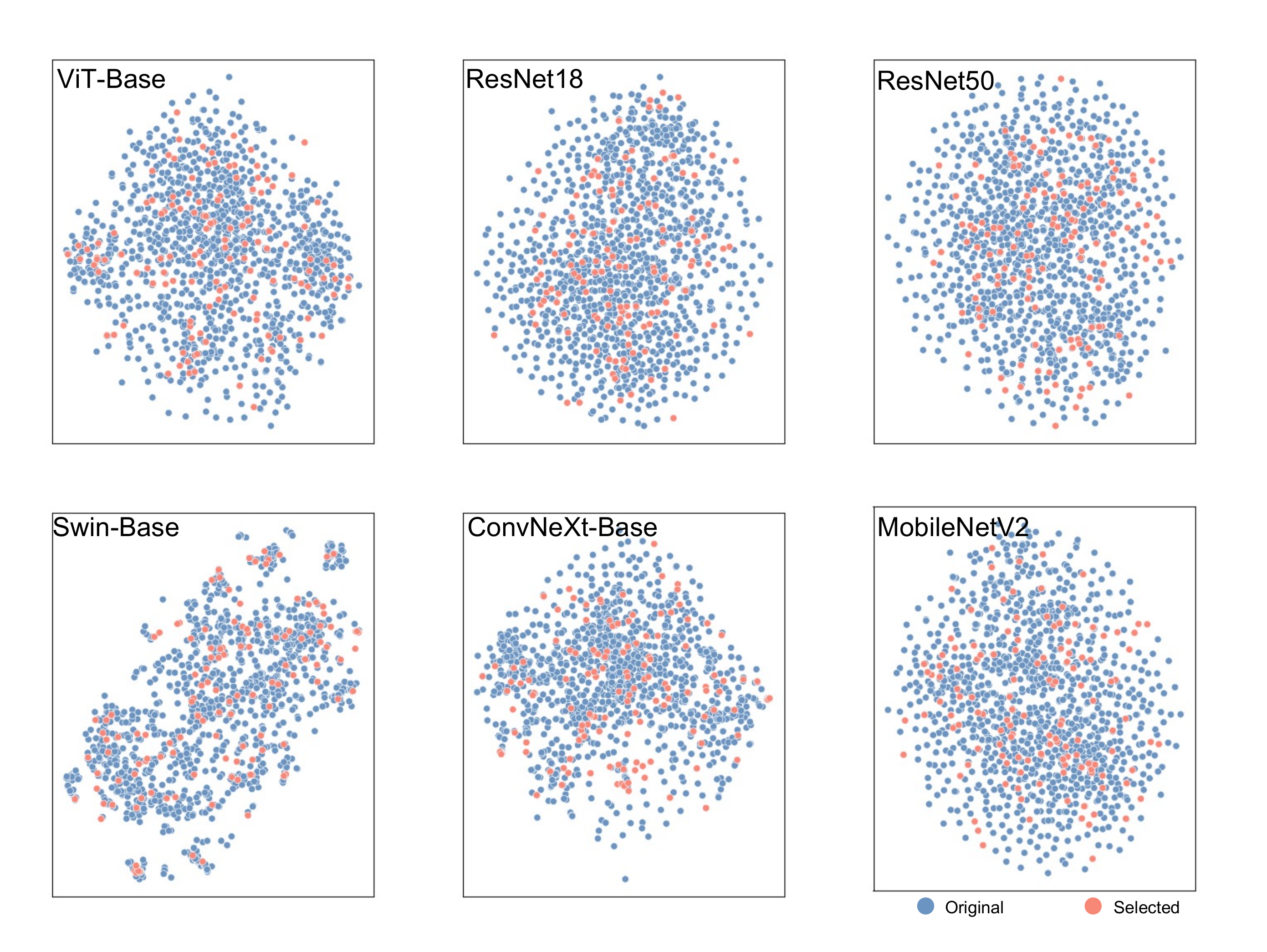}
    \caption{Cross-architecture visualization of the feature distributions among the dataset generated by DQ on ViT-Base on ImageNet dataset tench class. }
    \label{fig:in_010_cross_arch}
\end{figure}

\paragraph{Cross-architecture generalization of DQ} 
We further present more feature distribution visualizations with different network architectures on ImageNet-1K in Fig.~\ref{fig:in_010_cross_arch}. 
The samples are originally selected by ResNet-18 and reconstructed with MAE. 
Each set contains 10\% of the total data. 
As shown, across all architectures, the generated compact set can effectively cover the whole data distribution, presenting significant cross-architecture generalization capability. 

\end{document}

\newcommand{\nameofmethod}{{{dataset quantization}}}
\title{Dataset Quantization}

\onecolumn
\maketitle
\section{Appendix}
% % 1. sota on CIFAR-10 and IN-1K
% 2. Visualization
% 3. Differences between Coreset and ours
% 4. more implementation details: pytorch cifar, timm, mmdet, mmseg details.
% 5. code
We present more explanations of the proposed \nameofmethod, experiment results and visualizations in this section. 
% In Sec. , we produce the difference comparisons between coreset selection methods and our DQ, cross-architecture feature distributions, and code of DQ.
% \input{./tables/tab-2}

% \input{./tables/table_in}
\subsection{Proof of Sec. 3.1}
\label{appendix:proof}
Given the whole dataset $\mathbf{D}$, $|\mathbf{D}| = M \gg 1$. $\forall{p} \in \mathbf{D}$, $f(p) \in \mathbb{R}^{m \times 1}$.
To make a simple proof, we assume $\frac{1}{M}\sum_{p \in \mathbf{D}}f(p) = 0$.

For $n = 0, 1, 2, \cdots, n, \cdots, N$, define set $\mathbf{S}_{n} \in \mathbf{D}$. And, we define $C_{1}(x)$ and $C_{2}(x)$ as follows,

\begin{equation}
    C_{1}(x) = \sum_{p \in \mathbf{S}_{1}^{n}}||f(p) -f(x)||_{2}^{2}; \quad \quad \quad C_{2}(x) = \sum_{p \in \mathbf{D} \backslash \mathbf{S}_{1}^{i}}||f(p) -f(x)||_{2}^{2}; 
\end{equation}

By the policy of GraphCut~\cite{iyer2021submodular}, it aims to maximize $C_{1}(x)$ and minimize $C_{2}(x)$ to select $\mathbf{S}_{1}$. We write it into a united target function to choose $x_{k+1}$ as,

\begin{equation}
    x_{k+1} \leftarrow \argmax_{x \in \mathbf{D} \backslash \mathbf{S}_{1}^{k}}(C_{1}(x) - C_{2}(x)).
\end{equation}
We initialize $\mathbf{S}_{1}^{1}$ using $\emptyset$ and $\mathbf{S}_{1}^{k+1} = \mathbf{S}_{1}^{k} \cup {x_{k+1}}$, where $k = 1, 2, \cdots, k, \cdots, K$, and $|\mathbf{S}_{1}^{k}| = k$.

\textbf{Claim:} (a). $\mathbf{S}_{1}^{1} = \emptyset$, $x_{1} = \argmin_{x \in \mathbf{D}}||x||_{2}^{2}$, \textit{i.e,} the closest point to \textbf{0} in $\mathbf{D}$

(b). $x_{k+1}$ is very close to set $\mathbf{S}_{1}^{k}$.

\textbf{Proof:} $\mathbf{S}_{1}^{1} = \emptyset$, so $C_{1}(x) = 0$.
\begin{align}
    C_{2}(x) &= \sum_{p \in \mathbf{D}}||f(p)-f(x)||_{2}^{2}\\ 
    &= M||f(x)||_{2}^{2} - 2(\sum_{p \in \mathbf{D}}f(p))^{\top}f(x) + \sum_{p \in \mathbf{D}}||f(p)||_{2}^{2} \\
&= M||f(x)||_{2}^{2} + \sum_{p \in \mathbf{D}}||f(p)||_{2}^{2},
\end{align}
where $\sum_{p \in \mathbf{D}}f(p) = 0$. 

Then, we have 
\begin{align}
  x_{1} &= \argmax_{x \in \mathbf{D}} -C_{2}(x)\\
   &= \argmin_{x \in \mathbf{D}}C_{2}(x)\\
   &= \argmin_{x \in \mathbf{D}}M||f(x)||_{2}^{2},
\end{align}

(b) Let $C_{1}(x_{k}) - C_{2}(x_{k}) = 2C_{1}(x_{k}) - (C_{2}(x_{k}) + C_{1}(x_{k}))$.
We have:
\begin{align}
    (C_{2}(x_{k}) + C_{1}(x_{k})) &=\sum_{p \in \mathbf{D} \backslash \mathbf{S}_{1}^{k}}||f(p) - f(x_{k})||_{2}^{2} + \sum_{p \in  \mathbf{S}_{1}^{k}}||f(p) - f(x_{k})||_{2}^{2} \\
    &= \sum_{p \in \mathbf{D}}||f(p)-f(x)||_{2}^{2} \\
    &= M||f(x)||_{2}^{2} + \sum_{p \in \mathbf{D}}||f(p)||_{2}^{2} \\
    &= M||f(x)||_{2}^{2} + Const.,
\end{align}
where `Const.' denotes constant number.

For $C_{1}(x_{k})$, we have
\begin{equation}
   C_{1}(x_{k}) = \sum_{p \in \mathbf{S}_{1}^{k}}||f(p) - f(x_{k})||_{2}^{2} = 
k||f(x_{k})||_{2}^{2} -2(\sum_{p \in \mathbf{S}_{1}^{k}}f(p)^{\top}f(x_{k}) + \sum_{p \in \mathbf{S_{1}^{k}}}||f(p)||_{2}^{2} 
\end{equation}
Define $Q_{k} = \frac{1}{k}\sum_{p \in \mathbf{S}_{1}^{k}}f(p)$ as the weighted center of $\mathbf{S}_{1}^{k}$.
Then, we can write the submodular gains function as follows,

\begin{align}
    P(x_{k}) &= 2C_{1}(x_{k}) - (C_{2}(x_{k}) + C_{1}(x_{k})) \\
    &=2k||f(x_{k})||_{2}^{2} - 4kQ_{k}^{\top}f(x_{k}) - M||f(x_{k})||_{2}^{2} + Const. \\
    &= (2k - M) ||f(x_{k}) - \frac{2kQ_{k}}{2k - 
    M}||_{2}^{2} + Const.
\end{align}

$x_{k+1}$ is selected as follows,

\begin{equation}
    x_{k+1} = \argmax_{x \in \mathbf{D} \backslash \mathbf{S}_{1}^{k}}P(x_{k}) = \argmax_{x \in \mathbf{D} \backslash \mathbf{S}_{1}^{k}}||f(x_{k}) - \frac{2kC_{k}}{2k-M}||_{2}^{2}.
\label{star_eqn}
\end{equation}

Let $\delta_{k} = \frac{2kC_{k}}{2k-M}$. We define radius $R_{1}^{k}$ of set $\mathbf{S_{1}^{k}}$ as,

\begin{equation}
    R_{1}^{k} = max_{p \in \mathbf{S}_{1}^{k}}||f(p)||_{2}.
\end{equation}
Therefore, $\forall{p} \in \mathbf{S}_{1}^{k}$,
$||f(p)||_{2}^{2} \leq (R_{1}^{k})^2$, which means $\mathbf{S}_{1}^{k}$ is included in a ball $\mathbf{B}_{k} = \{p|||f(p)||_{2}^{2} \leq (R_{1}^{k})^2\}$.
% $\mathbf{B}_{k} = {p|||y||_{2}^{2} \leq R_{k}^2}$.
Note that,
\begin{align}
   ||\delta_k||_{2}^{2} &= (2k/2k-M)^2||Q_{k}||_{2}^{2} \\
   &= (\frac{2k}{2k-M})^{2}||\frac{1}{k}\sum_{p \in \mathbf{S}_{1}^{k}}f(p)||_{2}^{2}\\
   &\leq (\frac{2k}{2k-M})^{2}\frac{1}{k}\sum_{p \in \mathbf{S}_{1}^{k}}||f(p)||_{2}^{2}\\
   &\leq (\frac{2k}{2k-M})^{2}(R_{1}^{k})^{2}.
\end{align}

$M \gg k$, so $||\delta_{k}||_{2}^{2} \leq (R_{1}^{k})^{2}$ and $\delta_{k} \in \mathbf{B}_{k}$
According to Eq. \ref{star_eqn}, $x_{k+1} = \argmin_{x \in \mathbf{D} \backslash \mathbf{S}_{1}^{k}}||x - \delta_{k}||_{2}^{2}$ is the closest point in $\mathbf{D} \backslash \mathbf{S}_{1}^{k}$ to $\delta_{k}$, which is in the ball $\mathbf{B}_{k}$. As $M \gg 1$, $f(x_{k+1})$ is very close to $\mathbf{B}_{k}$, and thus to $\mathbf{S}_{1}^{k}$.

By the proof, GraphCut cannot guarantee the samples diversity under small data keep ratio.
Our DQ recursively select samples from $\mathbf{D}$, as the total number of $\mathbf{D}$ reduces, the radius of the ball $\mathbf{B_k}$ will be extended. Therefore the sample diversity is higher than GraphCut method.

\subsection{Source Code}
We have submitted the source code as the supplementary materials in a zipped file named as `DQ.zip' for reproduction. A README file is also included for the instructions fo running the code. We will make it public after the submission period.

\subsection{Details of Patch Dropping and Reconstruction}
\label{details_patch_attention}
%new version
As pointed out in Masked Auto-Encoder (MAE) ~\cite{he2022masked}, with a pre-trained decoder, some image patches can be dropped without affecting the reconstruction quality of the image. Motivated by it, we propose to reduce the number of pixels utilized for describing each image.
% RES can be regarded as the first step to select informative samples. 
% There still exists redundancy within each sample, such as redundant background or corrupted objects.
% Recent works ~\cite{} have analysed there exists redundancy information within an image, such as meaningless background or corrupted objects.
% There naturally raise a question: Can we quantize patches of a single image based on their informativeness?
% Our proposed RES can effectively reduce the redundant samples of dataset, how to reduce redundancy within each sample?
% However, the patches are the minimum description length for dataset.
% Specifically, we design a quantization method within each image, named as image quantizer (IQ). 
Specifically, as shown in pipeline, given an image $x$,
%target to get an importance score for each 
we first feed it into a pretrained feature extractor (ResNet-18 ~\cite{he2016deep}) 
%\JS{\emph{e.g.}?, what feature extractor used?} 
to obtain the last feature map $\mathcal{M}$ and a prediction score $y^c$ of the image class $c$. 
%The score of each pixel in the image $x$ with respect to the 
% predic
% calculate the gradient values of each pixel in the last feature map by GradCAM++ ~\cite{aditya1710grad}, which can be formulated as follows,
A group of attention scores is then calculated with the gradient values of each pixel in the last feature map following GradCAM++ ~\cite{aditya1710grad}:
% \JS{the following equation is identical with the one in this reference. change to some different symbols and adapt to our overall contents}:
% \begin{equation}
%  a_k^c = \sum_{i}\sum_{j}\alpha_{ij}^{kc}.relu(\frac{\partial Y^c}{\partial A_{ij}^k}), \label{weighted version}
% \end{equation}
\begin{equation}
 a^c = \sum_{i,j}\left[\frac{\frac{\partial ^2y^c}{(\partial \mathcal{M}_{ij})^2}}{2 \frac{\partial ^2y^c}{(\partial \mathcal{M}_{ij})^2}\! + \! \sum_{m,n} \mathcal{M}_{mn}\{\frac{\partial ^3y^c}{(\partial \mathcal{M}_{ij})^3}\}}\right] \mathrm{ReLU}\left(\frac{\partial y^c}{\partial \mathcal{M}_{ij}}\right),
\end{equation}
where $a^c$ is the attention scores for each pixel w.r.t.\ class $c$, $\mathrm{ReLU}$ is the Rectified Linear Unit activation function, and ($i$, $j$) and ($m$, $n$) are iterators over the feature map $A$.
%, $\alpha_{ij}^{kc}$ denotes the gradient weights.
% Here, we default to calculate the last convolutional layer's activation map of ResNet-18, so the size of $a^c_k$ is $7 \times 7$.
The pixel-wise attention score $a^c$ is upsampled to fully cover the original input image.
%\JS{this is not clear. how to interpolate?}.
In order to integrate the attention information 
%\JS{refer to?} 
into image patches, we unify the attention scores of the corresponding pixels of a patch by their average value to generate the patch-wise importance scores $p^c_{\cdot}$ as follows,
\begin{equation}
p^c_k=\frac{1}{hw}\sum_{i=h_k}^{h_k+h}\sum_{j=w_k}^{w_k+w}a^c(i,j),
\end{equation}
where $h_k$ and $w_k$ are the coordinates of the upper left corner of the patch $k$, and $h$ and $w$ are the height and width of image patches. 
% In our method, we further calculate the attention score for each patch, respectively.
% A patch-wise attention score is obtained by averaging pixel-wise attention scores within each patch.
% Then, we sort the patch-wise attention scores by descending order.
According to the patch-wise attention scores, we drop a percentage of $\theta$ non-informative patches with smallest attention scores to further save the storage cost.
At the training stage, we employ a strong pre-trained MAE decoder to reconstruct the dropped patches and the original images.
% \JS{need to explain how to use DQ processed dataset for model training}

\subsection{Robustness Evaluation}
\begin{figure*}[t]
\centering
    \begin{subfigure}{0.19\textwidth}
    \includegraphics[width=\textwidth,height=0.9\textwidth]{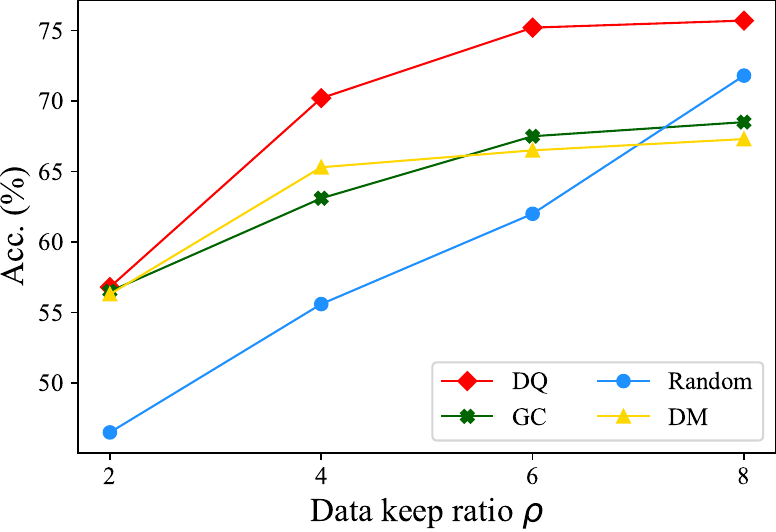}
    \caption{Level 1}
    \label{fig:5a}
    \end{subfigure}
    \begin{subfigure}{0.19\textwidth}
    \includegraphics[width=\textwidth,height=0.9\textwidth]{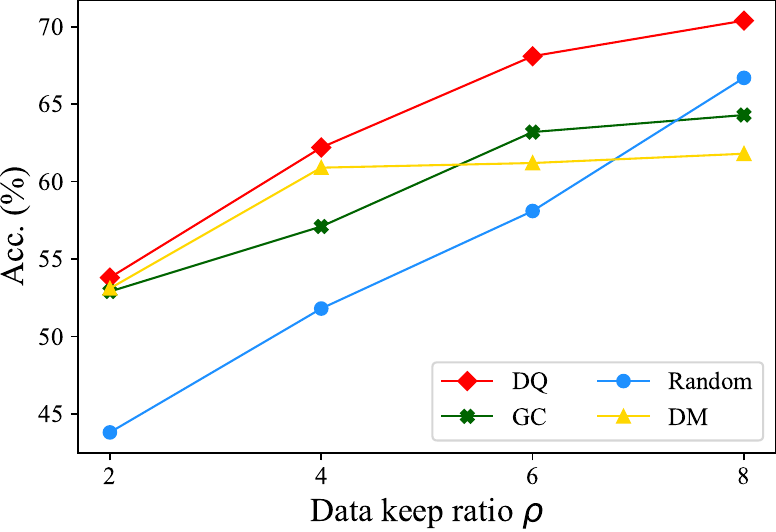}
    \caption{Level 2}
    \label{fig:5b}
    \end{subfigure}
    \begin{subfigure}{0.19\textwidth}
    \includegraphics[width=\textwidth,height=0.9\textwidth]{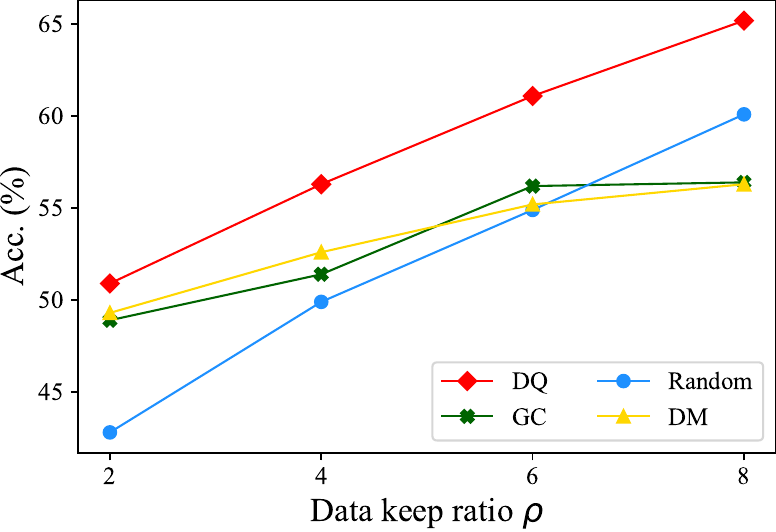}
    \caption{Level 3}
    \label{fig:5c}
    \end{subfigure}
    \begin{subfigure}{0.19\textwidth}
    \includegraphics[width=\textwidth,height=0.9\textwidth]{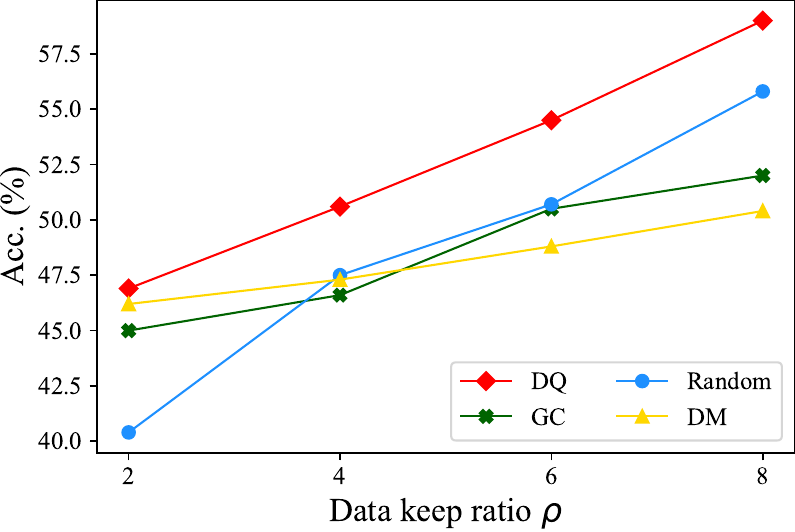}
    \caption{Level 4}
    \label{fig:5d}
    \end{subfigure}
    \begin{subfigure}{0.19\textwidth}
    \includegraphics[width=\textwidth,height=0.9\textwidth]{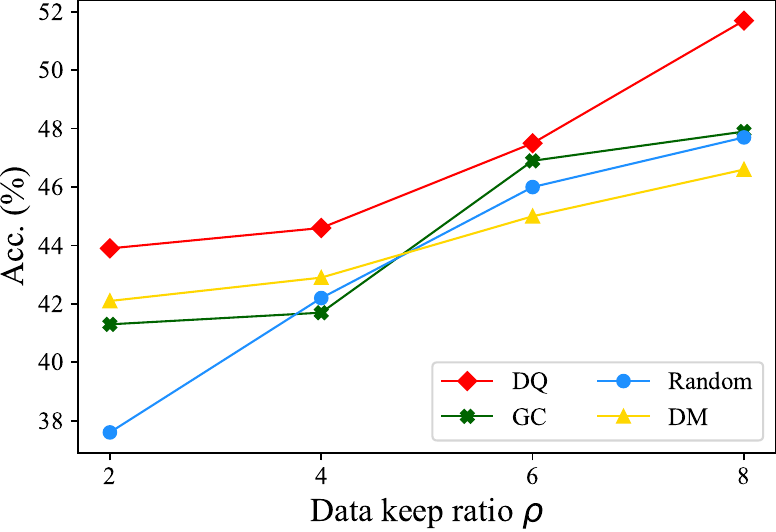}
    \caption{Level 5}
    \label{fig:5e}
    \end{subfigure}
    \caption{Comparisons of the robustness of trained models via DQ, GC and Random selection on CIFAR10-C dataset.} 
\label{fig:fig_robust}
\end{figure*}

We show the overall robustness evaluation in our paper. Here, we report the detailed results at different corruption levels in Tab.~\ref{tab:rbst_1} ~\ref{tab:rbst_2} ~\ref{tab:rbst_3} ~\ref{tab:rbst_4}, and Fig.~\ref{fig:fig_robust}. Our proposed DQ achieves state-of-the-art results in all cases.
\label{rbst_appendix}
\input{./tables/robustness}
\input{./tables/robustness_004}
\input{./tables/robustness_006}
\input{./tables/robustness_008}

\subsection{Differences between coreset selection and \nameofmethod}

\begin{figure}[t]
    \centering
    \includegraphics[width=\textwidth]{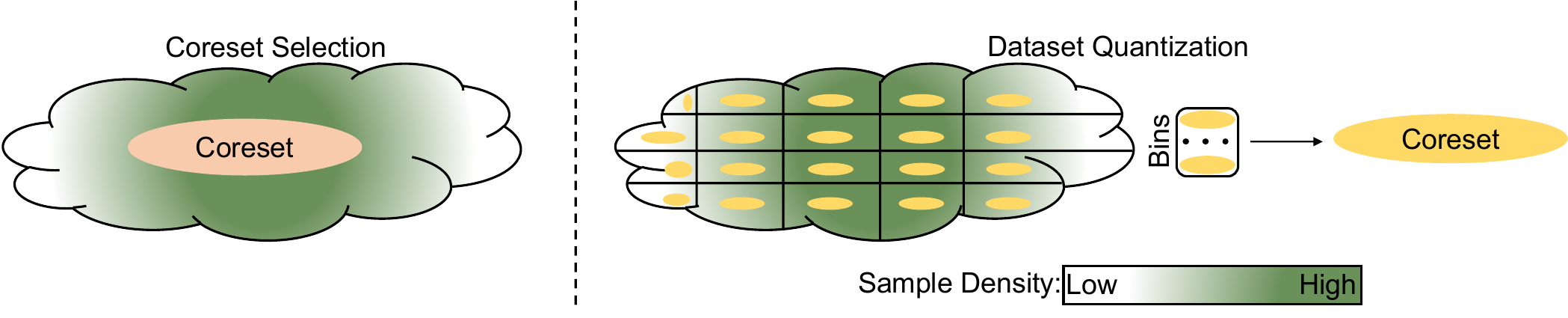}
    \caption{Differences between coreset selection methods and our dataset quantization.}
    \label{fig:coreset_comp}
\end{figure}

\begin{figure}[t]
    \centering
    \includegraphics[width=\textwidth]{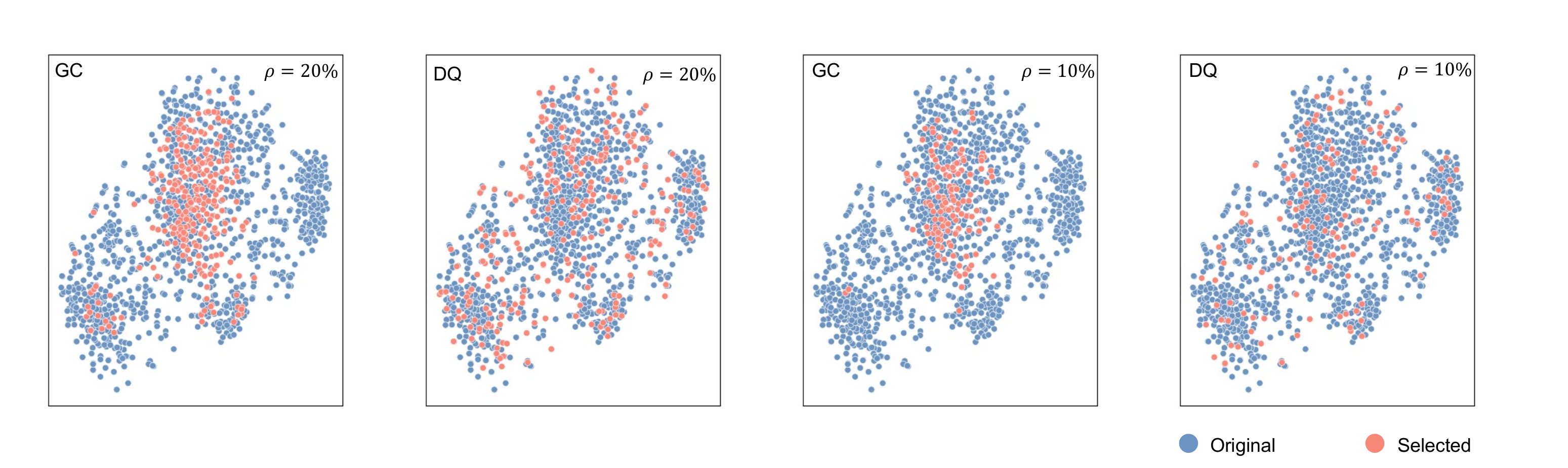}
    \caption{Visualization of the feature distributions among data selected by GraphCut and SQ.}
    \label{fig:vis_gc_dq}
\end{figure}

\paragraph{Coreset VS DQ} We here give more detailed explanations on the difference between the coreset selection methods and our proposed \nameofmethod. As shown in Fig. \ref{fig:coreset_comp}, the coreset selection only select one subset from the full data distribution. This practice will suffer from a selection bias, resulting in selection results with limited diversity. Besides, when the the size of the selected subset is small, it will suffer a large selection variance. Differently, \nameofmethod~first divides the full distribution into non-overlapping bins and then sampling from each bin uniformaly. As a result, the sampled data could maximally preserve the original data distribution. To verify this, we use GraphCut ~\cite{iyer2021submodular} as a representation of the coreset based method and 10\% and 20\% data from 
ImageNet dataset and compare the results with the data distribution sampled with \nameofmethod. We use a pre-trained ResNet-18 model to extract the features of the data and then visualize the extracted data via t-SNE. The results are shown in Fig. \ref{fig:vis_gc_dq}. It is clearly observed that the data sampled via \nameofmethod~ do capture a more diverse distribution.

\clearpage

\begin{figure}[t]
    \centering
    \includegraphics[width=.95\textwidth]{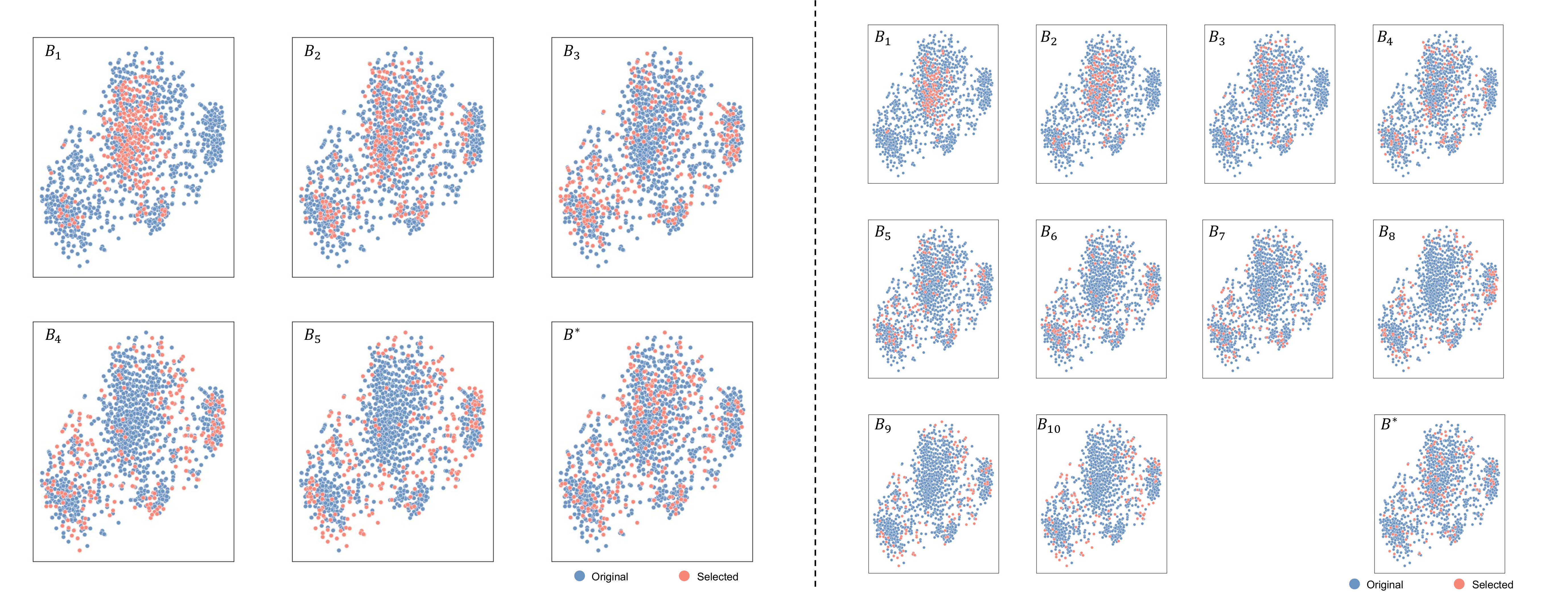}
    \caption{Visualization of the feature distributions among data selected in each bin and the final output of SQ on ImageNet dataset tench class. The bin number $N$ and the data keep ratio $\rho$ are set as (5, 20), (10, 10), respectively for the left and right column. }
    \label{fig:in_010_020_tsne}
\end{figure}

\paragraph{Bin diversity of DQ} To dig deeper for the reason why DQ can better preserved the data distribution. We use the same visualization method as aforementioned for the data contained within each bin. The results are shown in Fig. \ref{fig:in_010_020_tsne}. Each bin contains 20\% of the total data in the left column and 10\% data in the right column. As shown, different bins are capturing different distributions. As a results, after sampling uniformly from each bin, the combined dataset enjoys a large diversity as well as representativeness over the whole data distribution. 
%As we increase the bin number, each bin contains a more diverse data distribution and the combined subset sampled from each bin are also more diverse as shown in Fig. \ref{fig:in_010_020_tsne}.

\clearpage

\begin{figure}[ht]
    \centering
    \includegraphics[width=\textwidth]{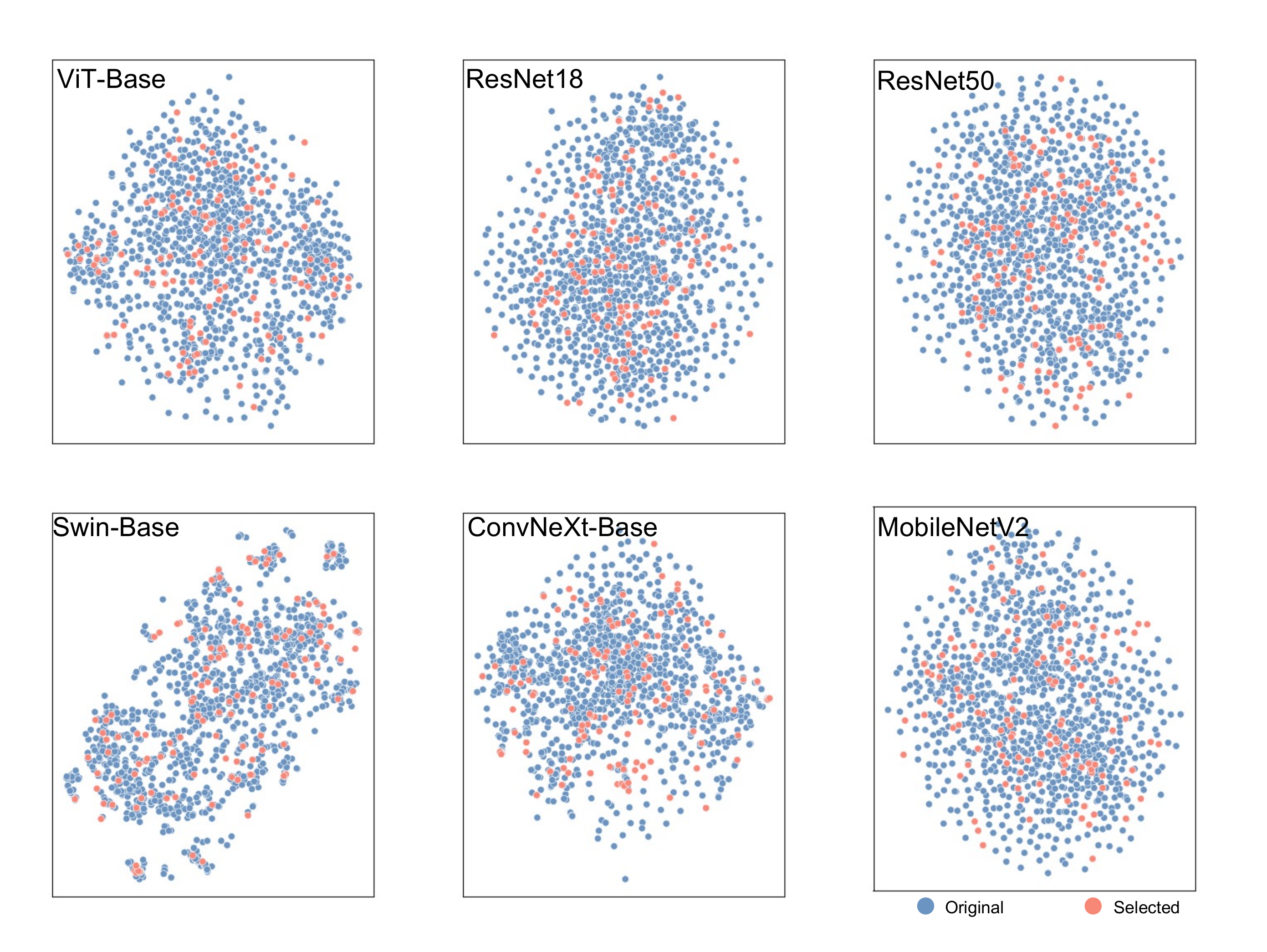}
    \caption{Cross-architecture visualiztaion of the feature distributions among the dataset generated by DQ on ViT-Base on ImageNet dataset tench class. }
    \label{fig:in_010_cross_arch}
\end{figure}

\paragraph{Cross-architecture generalization of DQ} 
We further present more feature distribution visualizations with different network architectures on ImageNet-1K in Fig.~\ref{fig:in_010_cross_arch}. 
The samples are originally selected by ResNet-18 and reconstructed with MAE. 
Each set contains 10\% of the total data. 
As shown, across all architectures, the generated compact set can effectively cover the whole data distribution, presenting significant cross-architecture generalization capability. 

{\small
\bibliographystyle{ieee_fullname}
\bibliography{iclr2023_conference}
}